\documentclass[preprint, review, 10pt]{elsarticle}

\usepackage{graphicx}
\usepackage{fullpage}
\usepackage{amssymb}
\usepackage{bbm}
\usepackage[normalem]{ulem}
\usepackage[dvipsnames]{xcolor}
\usepackage{comment}
\definecolor{mypink}{rgb}{0.858, 0.188, 0.478}

\usepackage{lineno}
\usepackage{verbatim}
\usepackage{xcolor}
\usepackage{eqnarray}
\usepackage{siunitx}
\usepackage[T1]{fontenc}
\usepackage{mathtools}
\RequirePackage{ifthen}
\RequirePackage{graphicx}
\RequirePackage{amsmath}
\RequirePackage{fancyhdr}
\usepackage{booktabs}
\usepackage{siunitx}
\usepackage{fancyhdr,graphicx,amsmath,amssymb,microtype}
\usepackage[ruled,vlined]{algorithm2e}
\SetKwRepeat{Do}{do}{while}
\usepackage{alltt}%
\SetAlFnt{\small}
\SetAlCapFnt{\normalsize}
\SetAlCapNameFnt{\normalsize}
\usepackage{algorithmic}
\algsetup{linenosize=\small}
\include{pythonlisting}
\usepackage{setspace}
\usepackage[thinc]{esdiff}
\usepackage{booktabs}
\usepackage{multirow}
\usepackage{glossaries}
\newacronym{bc}{BC}{boundary condition}
\usepackage{derivative}
\usepackage{mathrsfs}
\usepackage[utf8]{inputenc}
\usepackage[english]{babel}

\newcommand{\inff}[1]{\underset{#1}{\operatorname{inf}}}
\definecolor{purp}{rgb}{0.4,0.2,0.8}

\usepackage{graphicx}
\usepackage{mathrsfs}
\usepackage{caption}
\usepackage{subcaption}
\usepackage{hyperref}
\usepackage{lineno}

\definecolor{custom-blue}{RGB}{3,69,173}
\hypersetup{
    colorlinks = true,
    urlcolor   = blue,
    citecolor  = custom-blue,
}

\newcommand*{\review}{\textcolor{black}}

\biboptions{sort&compress}

\begin{document}

\begin{frontmatter}


\title{A survey of unsupervised learning methods for high-dimensional uncertainty quantification in black-box-type problems}


\author[1]{Katiana Kontolati}
\ead{kontolati@jhu.edu}
\author[2,3,4]{Dimitrios Loukrezis}
\ead{dimitrios.loukrezis@\{tu-darmstadt.de, siemens.com\}}
\author[1]{Dimitrios G. Giovanis}
\ead{dgiovan1@jhu.edu}
\author[1]{Lohit Vandanapu}
\ead{lvandan2@jhu.edu}
\author[1]{Michael D. Shields\corref{cor1}}
\ead{michael.shields@jhu.edu}

\address[1]{Department of Civil \& Systems Engineering, Johns Hopkins University, Baltimore MD, USA}

\address[2]{Institute for Accelerator Science and
Electromagnetic Fields, Technische Universität Darmstadt, Darmstadt, Germany}

\address[3]{Centre for Computational Engineering, Technische Universität Darmstadt, Darmstadt, Germany}

\address[4]{Siemens AG, Technology, Munich, Germany}

\cortext[cor1]{Corresponding author.}

\begin{abstract}
Constructing surrogate models for uncertainty quantification (UQ) on complex partial differential equations (PDEs) having inherently high-dimensional \review{$\mathcal{O}(10^n)$, $n \geq 2$}, stochastic inputs (e.g., forcing terms, boundary conditions, initial conditions) poses tremendous challenges. The ``curse of dimensionality'' can be addressed with suitable unsupervised learning techniques used as a pre-processing tool to encode inputs onto lower-dimensional subspaces while retaining its structural information and meaningful properties. In this work, we review and investigate thirteen dimension reduction methods including linear and nonlinear, spectral, blind source separation, convex and non-convex methods and utilize the resulting embeddings to construct a mapping to quantities of interest via polynomial chaos expansions (PCE). We refer to the general proposed approach as \textit{manifold PCE} (m-PCE), where manifold corresponds to the latent space resulting from any of the studied dimension reduction methods. To investigate the capabilities and limitations of these methods we conduct numerical tests for three physics-based systems (treated as black-boxes) having high-dimensional stochastic inputs of varying complexity modeled as both Gaussian and non-Gaussian random fields to investigate the effect of the intrinsic dimensionality of input data. We demonstrate both the advantages and limitations of the unsupervised learning methods and we conclude that a suitable m-PCE model provides a cost-effective approach compared to alternative algorithms proposed in the literature, including recently proposed expensive deep neural network-based surrogates and can be readily applied for high-dimensional UQ in stochastic PDEs.

\end{abstract}


\end{frontmatter}





\section{Introduction}
\label{S:Introduction}

In real-world physics and engineering applications, uncertainty quantification (UQ) plays a pivotal role, as it allows for the characterization, quantification and reduction of uncertainties associated with model predictions. Exploring complex systems under the presence of both aleatoric (inherent stochasticity) and epistemic (lack of knowledge) uncertainties \cite{der2009aleatory} allows for reliability analysis, risk modelling and design optimization \cite{sullivan2015introduction}. State-of-the-art computational models involve a large number of spatially and temporally variable/uncertain parameters and inputs that greatly affect the overall response of the model. The process of quantifying the variability and quality of model predictions that result from these input uncertainties is known as uncertainty propagation or forward UQ. 

Monte Carlo (MC) methods, invented in the late 1940s at Los Alamos National Laboratory \cite{metropolis1987beginning, liu2001monte}, involve repeated random sampling to propagate uncertainties and compute integral quantities of interest (QoI), e.g. moments of the model output. MC-based methods are simple to implement yet impractical for resource-demanding models with long runtimes. Modern MC approaches benefit significantly from variance reduction methods, such as stratified / Latin hypercube sampling \cite{mckay2000comparison, helton2003latin, shields2016generalization} and importance sampling \cite{tokdar2010importance}, that reduce the number of required random samples. But these methods often do not scale well with dimension and may still require very large sample sizes. More advanced MC methods have been also developed which leverage added information from related models using control variates \cite{gorodetsky2020generalized} and related concepts, such as the multi-fidelity Monte Carlo (MFMC) \cite{peherstorfer2016optimal, peherstorfer2018multifidelity} and the multi-level Monte Carlo (MLMC) \cite{giles2015multilevel, krumscheid2020quantifying, scheichl2017quasi} methods, to reduce computational cost. 

\subsection{Surrogate modeling}

An alternative, and often more efficient, approach is to construct a surrogate model, also known as an emulator, response surface, or metamodel \cite{sudret2017surrogate, bhosekar2018advances}, which approximates the input-output mapping by an analytical or machine learned function that is fast to compute and accurate. Common methods include Gaussian process (GP) regression or Kriging \cite{rasmussen2003gaussian, bilionis2012multi, chen2015uncertainty, tripathy2016gaussian, raissi2018numerical, radaideh2020surrogate, giovanis2020data, kontolati2021manifold2}, polynomial chaos expansions (PCEs) \cite{ghanem1990polynomial, xiu2002wiener, witteveen2006modeling, oladyshkin2012data, zheng2015adaptive, kontolati2021manifold}, and artificial neural networks (ANNs) \cite{psichogios1992hybrid, lagaris1998artificial, zhu2018bayesian, zhang2019quantifying, winovich2019convpde, yang2019adversarial, raissi2019physics, olivier2021bayesian, lu2021learning, goswami2021physics, gao2021wasserstein, kontolati2022influence}, among others. The surrogate modeling framework can be very efficient when the dimension of the input is modest. However, for models with very high input dimension, surrogate model construction can be intractable; a consequence of the \textit{curse of dimensionality} \cite{keogh2017curse} which derives from the exponential growth in the volume of the input space with increasing dimension. 

 Perhaps the most widely-used surrogate modeling technique for UQ, PCE also suffers from the curse of dimensionality as the basis of orthogonal polynomials grows rapidly with increasing input dimension. To reduce the dimension of PCEs, several sparse-grid/basis adaptation approaches have been developed for the construction of reduced in size PCEs, and an excellent review of these methods is presented by L{\"u}then et al.\ \cite{luthen2021sparse, luthen2020benchmark}. Tiripaddy and Ghanem \cite{tipireddy2014basis} proposed a method for transforming the probabilistic input space to a suitable Gaussian one that has been successfully applied to practical engineering problems \cite{thimmisetty2017homogeneous, ghauch2019integrated}. The method was also extended from scalar QoIs to random fields by Tsilifis and Ghanem \cite{tsilifis2017reduced} and was also combined with compressive sensing algorithms \cite{tsilifis2019compressive, kougioumtzoglou2020sparse}, which results in a reduced PCE approximation with optimal sparsity. Hampton and Doostan \cite{hampton2018basis} propose a basis adaptive sample efficient PCE method, also based on compressive sensing. Recently, Zeng et al.\ proposed an extension of the original basis adaptation approach to accelerate convergence \cite{zeng2021accelerated}. A partial least squares (PLS) basis adaptation approach has been proposed by Papaioannou et al.\ \cite{papaioannou2019pls}, which aims to identify the directions with the largest predictive significance in the PCE representation. A similar approach was developed by Zhou et al.\ \cite{zhou2020surrogate}, where a variant version of the PLS algorithm called sparse partial least square (SPLS) is employed. Ehre et al.\ \cite{ehre2021sequential}, proposed an extension of PLS-PCE with sequential active learning for reliability analysis problems. Other approaches to build sparse PCE models include the use of adaptive regression-based algorithms \cite{blatman2010adaptive}, \cite{pan2017sliced}, adaptive least-squares \cite{loukrezis2020robust}, $\ell$-1 minimization to recover important PCE coefficients \cite{jakeman2017generalized,salehi2018efficient,guo2018gradient,rauhut2012sparse}, least angle regression (LAR) \cite{hesterberg2008least, blatman2011adaptive, man2021generalized}, least absolute shrinkage and selection operator modified least angle regression algorithm (LASSO-LAR) \cite{meng2017efficient, meng2019efficient} and the maximum entropy principle \cite{he2020adaptive} for feature selection. The above mentioned methods allow for the construction of sparse PCE surrogates by reducing the size of the polynomial basis, however this size is dependent on both the order of polynomials and the input space dimension. Therefore constructing a PCE surrogate for very large input space dimensions still remains an unresolved issue.

\subsection{Dimension reduction for high-dimensional UQ}

To overcome this bottleneck, dimension reduction (DR) techniques can be employed to reduce high-dimensional input spaces to a small number of expressive components. DR refers to the process of reducing the number of features (also dimensions or attributes) in a dataset while preserving as much information in the original dataset as possible. Reducing data dimensions helps to avoid overfitting and allows for straightforward data visualization and feature selection by removing noise and redundant features. Furthermore, lower data dimensionality typically implies reduced training time, fewer computational resources and enhanced model performance. Ideally, the dimension of the compressed representation should correspond to the intrinsic dimension of the data. The intrinsic dimension refers to the minimum number of features or attributes (dimensions) needed to accurately account for the observed properties of the data and can also be used as a measure of complexity of the data \cite{van2009dimensionality}. 

DR-based high-dimensional UQ generally follows a two-step approach. The first step involves the discovery of a lower-dimensional space/manifold on which high-dimensional inputs live (an unsupervised learning task), and the second step is surrogate modeling on the reduced dimension (a supervised learning task). Numerous examples of this approach can be found in the literature. Damianou and Lawrence \cite{damianou2013deep} construct GP surrogates through a fully Bayesian framework that performs well even if data is scarce. Calandra et al.\  \cite{calandra2016manifold} propose the ``manifold GP'' by first transforming the data onto a feature space and constructing a GP model from the feature space to the observed space. Another class of papers focuses on active subspaces (AS) \cite{constantine2015active}, which identifies the projection subspace of highest variability through a decomposition of the covariance matrix of the gradient of the model. Tripathy et al.\ \cite{tripathy2016gaussian}, employ AS and construct a GP-based mapping to the output space. Additional AS-based methods for characterizing the effects of high-dimensional input uncertainties can be found in \cite{constantine2014computing, constantine2015exploiting}. Although powerful as a DR method, AS can be computationally prohibitive on complex models where the gradient is hard to obtain. A framework based on kernel-PCA in conjunction with Kriging or PCE is proposed by Lataniotis et al.\ \cite{lataniotis2020extending}. \review{Low-rank tensor-based schemes have been also proposed to to address the issue of high-dimensional input \cite{doostan2013non,konakli2016reliability, bigoni2016spectral,gorodetsky2018gradient,he2020high}}.

Another class of methods involve the use of Deep Neural Networks (DNNs) which provide a powerful tool to learn the latent input representation automatically by supervising model response. Hinton and Salakhutdinov \cite{hinton2006reducing} proposed the use of an NN with a small central layer, also known as an \textit{autonencoder}, to efficiently reduce the number of features for high-dimensional input vectors which can be concurrently used for surrogate modeling. GPs can be used in conjunction with DNNs as in the work of Huang et al.\ \cite{huang2015scalable}, where a scalable GP model (DNN-GP) for regression is developed by applying a deep neural network as a feature-mapping function. Li et al.\ \cite{li2020deep}, propose a similar approach where a DNN and a GP are employed to tackle problems in reliability analysis. Alternative NN architectures that have been developed for high-dimensional UQ involve convolutional neural networks (CNNs) by Khoo et al.\ \cite{khoo2021solving}, and convolutional autoencoders by Boncoraglio and Farhat \cite{boncoraglio2021active}. A Bayesian approach was developed by Zhu and Zabaras \cite{zhu2018bayesian}, by proposing Bayesian deep convolutional encoder–decoder networks for UQ in problems governed by stochastic PDEs when limited training data is available. To address the issue of high-input dimensionality Zhu et al.\ \cite{zhu2019physics} followed a physics-informed deep learning approach by incorporating the governing equations to the loss function of the trained network. Despite showing good performance, this approach is associated with several difficulties related to the construction of the problem-specific loss function. A DNN-based method, which can be thought of as the nonlinear generalization of the AS method was proposed by Tripathy and Bilionis \cite{tripathy2018deep} to recover a low-dimensional manifold and construct a lower-dimensional DNN surrogate. A combination of DNNs and arbitrary PC (aPC) was developed by
Zhang et al.\ \cite{zhang2019quantifying}, to endow DNNs with uncertainty quantification for both parametric uncertainty and approximation uncertainties and allow for direct UQ in multi-dimensional stochastic PDEs. \review{A novel approach based on physics-informed normalizing flows was proposed by Guo et al.\ \cite{guo2022normalizing} for solving forward and inverse stochastic PDE problems.} More recently, a neural operator model known as the DeepONet has been proposed by Lu et al.\ \cite{lu2021learning}, to learn nonlinear operators and allow for high-dimensional UQ in complex PDEs. Although DNN-based methods are powerful and expressive function approximators, capable of capturing the complex nonlinear mapping between high-dimensional input and output, some issues remain related to the task of selecting the optimal network structure and tuning the values of the hyperparameters. Furthermore, training of such models can be prohibitively expensive. These challenges often render DNNs unfavorable from the UQ community.

\subsection{Motivation and objectives of this study}

Several studies in the literature have explored and compared DR techniques as a pre-processing step for feature transformation, selection and encoding of high-dimensional datasets \cite{van2009dimensionality, gao2017learning, chandrashekar2014survey, cunningham2015linear, gite2018comparative, sellami2018comparative, konstorum2018comparative, ayesha2020overview, alkhayrat2020comparative}. Motivated by the lack of a systematic comparison of such techniques applied for high-dimensional UQ in physics-based problems, this paper presents a comparative study of the most important linear and nonlinear DR techniques in this context. Specifically, we investigate the following thirteen methods: (1) principal component analysis (PCA), (2) kernel PCA, (3) independent component analysis (ICA), (4) non-negative matrix factorization (NMF), (4) isometric feature mapping (Isomap), (6) diffusion maps (DMAP), (7) locally linear embedding (LLE), (8) Laplacian eigenmaps or spectral embedding (SE), (9) Gaussian random projection (GRP), (10) sparse random projection (SRP), (11) t-distributed stochastic neighbor embedding (t-SNE), (12) neural autoencoders (AE) and (13) Wasserstein autoencoders (WAE). The objective of this paper is to investigate which classes of DR techniques perform best for the construction of surrogate models with compressed inputs. Furthermore, we aim to identify the limitations of these techniques. For the construction of surrogate models we use PCE, a native UQ method which performs well in the small data regime and allows for direct moment estimation and sensitivity analysis. Herein, we will refer to the proposed general model as \textit{manifold PCE} (m-PCE), where `manifold' refers to the reduced input space provided by any of the studied DR methods. The performance of each approach can be evaluated both in terms of the quality of the data compression, the accuracy of the surrogate, and the computational expense of the training. The quality of the data reduction is not necessarily correlated to the efficiency of the surrogate model \cite{lataniotis2019data}. However, within the context of UQ and predictive analysis, the ultimate goal is to achieve an optimal surrogate response. Thus, we will hereafter maintain a focus on the latter. 

We aim to cover a broad range of applications, thus, we test the performance of the proposed framework using three illustrative test cases of increasing complexity. Three physics-based models are considered, all of which are treated as black-boxes. High-dimensional stochastic inputs are represented as either Gaussian or non-Gaussian stochastic processes generated with the Karhunen Lo\'eve expansion (KLE) and the spectral representation method (SRM). These methods enable the generation of input fields of varying intrinsic dimensionality, thus allowing the exploration of the proposed approach for various degrees of difficulty. The proposed framework is used for predictive analysis, uncertainty propagation, and moment estimation. All surrogate modeling tasks were implemented in UQpy (Uncertainty Quantification with python), a general-purpose open-source software for modeling uncertainty in physical and mathematical systems \cite{olivier2020uqpy}. All manifold learning tasks were carried out using either scikit-learn 1.0 \cite{scikit-learn} or UQpy. 

The organization of this paper is as follows. In Section \ref{S:Methods}, we present the mathematical framework behind the linear and nonlinear DR techniques studied in this work. In Section \ref{S:approach}, we introduce the approach we employ for the construction of PCE surrogates with reduced input parameters for high-dimensional UQ. In Section \ref{S:Examples}, we provide a detailed study of the performance and predictive accuracy of the various methods by implementing them to three examples. The paper is concluded with a brief discussion in Section \ref{S:Discussion}.

\section{Dimension reduction methods}
\label{S:Methods}


In this work, we specifically apply 13 different methods from four general categories of dimension reduction: 1. Linear methods, 2. Nonlinear spectral methods, 3. Blind source separation methods, and 4. Non-convex methods. These methods are briefly introduced in the sections below with the following notational disclaimer. Consistency in notation is very challenging across the thirteen methods presented. Throughout the paper $\mathbf{X} = \{ \mathbf{x}_1,..., \mathbf{x}_N\}$, where $\mathbf{x}_i \in \mathbb{R}^{D}$ represents the input data realizations, $D$ is the dimensionality of the original data and $d$ is the dimensionality of the reduced space. Matrices are uppercase bold (however the opposite is not necessarily true, for example in k-PCA $\mathbf{\Phi}$ represents a mapping and not a matrix), vectors are lowercase bold, matrix/vector components (e.g., $k_{ij}$) and scalars (e.g., $D$ or $N$) are not bold, but can be either uppercase or lowercase. Notationally, each method should be considered stand on its own and does not depend on the notation of any other method, and while we aim to avoid notational redundancies some re-use of notation is unavoidable across different methods.

\subsection{Linear methods}

Linear methods apply a linear operation to the input data to transform the data onto a new rotated and/or stretched basis in Euclidean space that has lower dimension. We consider the two linear DR approaches: Principal Component Analysis (PCA) and Random Projection (RP). 

\subsubsection{Principal Component Analysis (PCA)}
\label{PCA}
Principal components analysis (PCA) is perhaps the most widely used DR technique. It is closely related to the singular value decomposition (SVD) and the (discrete) Karhunen-Lo\'eve expansion \cite{fodor2002survey, de2000multilinear, huang2001convergence} such that the three terms are sometimes used interchangeably. PCA aims to identify the set of orthogonal directions along which the variance of
the data set is the highest, or the so-called principal directions. Assuming the distribution of $N$ data points $\mathbf{X} = \{ \mathbf{x}_1,..., \mathbf{x}_N\}$, where $\mathbf{x}_i \in \mathbb{R}^{D}$, the sample covariance matrix is computed as
\begin{equation}
\label{eq:cov}
    \mathbf{\Sigma}_{D\times D} = \frac{1}{N}\bar{\mathbf{X}}\bar{\mathbf{X}}^{\text{T}},
\end{equation}
where $\bar{\mathbf{X}}$ denotes the centered (zero-mean) data. Performing a spectral decomposition of the matrix $\mathbf{\Sigma}$ gives
\begin{equation}
\label{eq:spectral}
    \mathbf{\Sigma} = \mathbf{W} \mathbf{\Lambda} \mathbf{W}^{\text{T}},
\end{equation}
where $\mathbf{W}$ is a $D \times d$ orthogonal matrix containing the eigenvectors and $\mathbf{\Lambda} = \text{diag}(\lambda_1,..,\lambda_d)$ is a diagonal matrix containing the ordered eigenvalues $\lambda_1 \le ... \le \lambda_d$ where $d$ is generally selected such that $\dfrac{\sum_{i=1}^d \lambda_i}{\sum_{i=1}^D \lambda_i}\ge \epsilon$ where $\epsilon$ represents the lowest acceptable  variance proportion. The principal components (PCs) are then given by the $d$ rows of the $d \times N$ matrix $\mathbf{S}$ given by
\begin{equation}
\label{eq:pcs}
    \mathbf{S} = \mathbf{W}^{\text{T}} \mathbf{X},
\end{equation}
where the above transformation maps a data vector $\mathbf{x}_i$ onto a new $d$-dimensional space spanned by $\mathbf{W}$ in which individual dimensions are linearly uncorrelated. The main drawback of PCA is the assumption that data are linearly separable and can be efficiently mapped onto linear hyperplanes. When this assumption fails, nonlinear DR methods need to be employed.

\subsubsection{Sparse and Gaussian Random Projection}
Random projection (RP) is a computationally simple linear DR method, in which the original high-dimensional dataset is projected onto a lower-dimensional subspace using a random matrix whose columns have unit lengths. RP has been shown to preserve the structural information of high-dimensional data and this has enabled a broad range of applications \cite{bingham2001random, goel2005face, cannings2017random, xu2017dppro}. Assuming $N$ realizations of the $D$-variate random vector $\mathbf{x} \in \mathbb{R}^D$, let the data matrix be $\mathbf{X}_{D \times N}$. RP projects the high-dimensional data to a $d$-dimensional subspace through the origin using a random matrix $\mathbf{R}_{d \times D}$ with unit-length columns as
\begin{equation}
\label{eq:random-projection}
    \mathbf{X}_{d \times N}^{\textit{RP}} = \mathbf{R}_{d \times D} \mathbf{X}_{D \times N}.
\end{equation}
This method is enabled by the Johnson-Lindenstrauss lemma, which states that if data points in a vector space are projected onto a randomly selected subspace of suitable dimension, then the pairwise distances between the points are approximately preserved. Proof of the J-L lemma can be found in \cite{dasgupta1999elementary}. Note that the vectors of $\mathbf{R}$ are not required to be orthogonal and that it has been shown that vectors having random directions are sufficiently close to orthogonal \cite{bingham2001random}. To evaluate the performance of the method, we can measure the similarity of two vectors in the reduced space. After the random projection the distance between two vectors $x_1, x_2$ is approximated by the Euclidean distance in the lower-dimensional space as
\begin{equation}
\label{eq:dist-reduced-space}
    \sqrt{(D/d)} \|\mathbf{R}x_1 - \mathbf{R}x_2 \|
\end{equation}
where the term $\sqrt{(D/d)}$, takes into account the reduction of the data dimensionality. The dimension $d$ can be selected based on the above value which the method aims to minimize.  

When the elements $r_{ij}$ of $\mathbf{R}$ are Gaussian-distributed, then the method is known as Gaussian random projection (GRP). Achlioptas \cite{achlioptas2001database} has proposed the use of sparse random matrices where the Gaussian distribution is replaced with the following simpler distribution
\begin{equation}
\label{eq:sparse-matrix}
r_{ij} = \sqrt{3} \left\{
        \begin{array}{lll}
            +1 & \quad \text{with probability} \hspace{3pt} \frac{1}{6}, \\
            0 & \quad \text{with probability} \hspace{3pt} \frac{2}{3}, \\
            -1 & \quad \text{with probability} \hspace{3pt} \frac{1}{6}.
        \end{array}
    \right.
\end{equation}
When the random matrix is distributed according to Eq.~\eqref{eq:sparse-matrix}, the method is known as sparse random projection (SRP) and often leads to computational savings. A novel tensor random projection (TRP) method was recently proposed \cite{sun2021tensor}, which requires substantially less memory than the existing matrix-based RP methods. Both the GRP and the SRP methods will be applied for DR in this study.

\subsection{Nonlinear spectral methods}

A set of widely-used approaches for DR are built from nonlinear spectral methods where low-dimensional embeddings are derived from the eigendecomposition of specially constructed matrices that express data similarity or affinity \cite{saul2006spectral}. Both kernel-based and graph-based methods are considered, where the latter methods construct a graph on the data whose nodes represent input patterns and whose edges represent neighborhood relations. An eigendecomposition of these graphs can reveal the low-dimensional structure of the intrinsic submanifold on which the high-dimensional data live.

\subsubsection{Kernel PCA}
Kernel PCA (k-PCA) \cite{scholkopf1997kernel, hoffmann2007kernel, mika1998kernel} extends standard PCA to nonlinear data distributions. Initially, data points are mapped into a higher-dimensional feature space $\mathcal{F}$, as
\begin{equation}
\label{eq:map}
    \mathbf{x}_i \rightarrow \mathbf{\Phi}(\mathbf{x}_i),
\end{equation}
where $\mathbf{\Phi}: \mathbb{R}^D \rightarrow \mathbb{R}^N$. Standard PCA is then performed in this higher-dimensional space. The PCs in $\mathcal{F}$ are theoretically obtained from the eigendecomposition of the sample covariance matrix. However, since $\mathbf{\Phi}$ creates linear independent vectors, there is no covariance on which to perform the decomposition. Instead, the above transformation is obtained implicitly through the use of a kernel function $k(\mathbf{x}_i, \mathbf{x}_j)$ which replaces the scalar product using the relation  $k(\mathbf{x}_i, \mathbf{x}_j) = \mathbf{\Phi}(\mathbf{x}_i)^{\text{T}} \mathbf{\Phi}(\mathbf{x}_j)$, thus $\mathbf{\Phi}$ is never calculated explicitly. This process is known as the `kernel trick' and allows the transformation of datasets in very high-dimensional spaces (even with infinite dimensionality), therefore allowing the encoding of highly nonlinear manifolds without the explicit knowledge of suitable feature functions \cite{bishop2006pattern}. We note that, because data in the featured space are not guaranteed to have zero-mean, the kernel matrix is centralized as
\begin{equation}
\label{eq:center}
    \mathbf{K}' = \mathbf{K} - \mathbf{1}_N \mathbf{K} - \mathbf{K} \mathbf{1}_N + \mathbf{1}_N \mathbf{K}, \mathbf{1}_N,
\end{equation}
where $\mathbf{1}_N$ represents an $N \times N$ matrix with values $1/N$ and $\mathbf{K}$ is the kernel matrix.

The k-PCA formulation does not allow the the PCs to be computed directly, but rather it computes the projections of the data onto the PCs. The projection of a mapped data point $\mathbf{\Phi}(\mathbf{x}_i)$ onto the $k$-th PC $\mathbf{W}^k$ in $\mathcal{F}$ computed as
\begin{equation}
\label{eq:kernel-pca}
    (\mathbf{W}^k)^{\text{T}} \cdot \mathbf{\Phi}(\mathbf{x}) = \bigg(\sum_{i=1}^{n}\alpha_i^k \mathbf{\Phi}(\mathbf{x}_i) \bigg)^{\text{T}} \mathbf{\Phi}(\mathbf{x}),
\end{equation}
where again $\mathbf{\Phi}(\mathbf{x}_i)^{\text{T}}\mathbf{\Phi}(\mathbf{x})$ is the dot product obtained from the entries of $\mathbf{K}'$. The coefficients $\alpha_i^k$, are determined by solving the eigenvalue problem
\begin{equation}
\label{eq:eigen}
    N \lambda \boldsymbol{\alpha} = \mathbf{K}' \boldsymbol{\alpha},
\end{equation}
where $\lambda$ and $\boldsymbol{\alpha}$ are the eigenvalues and eigenvectors of $\mathbf{K}'$ respectively and $N$ is the number of data points. Similar to standard PCA, the reduced dimension $d$ is based on the lowest acceptable variance proportion.  

Application of the method is dependent on the selection of a positive semi-definite kernel function, of which there are many options. Examples include the linear kernel $k(\mathbf{x}_i, \mathbf{x}_j) = \mathbf{x}_i^{\text{T}} \mathbf{x}_j$ which results in the standard PCA and the polynomial kernel $k(\mathbf{x}_i, \mathbf{x}_j) = (\mathbf{x}_i^{\text{T}} \mathbf{x}_j+c)^p$, $c \in \mathbb{R}_{\geq 0}$, $p \in \mathbb{Z}_{\geq 0}$. The most widely-used kernel is the Gaussian defined by
\begin{equation}
\label{eq:gaussian}
    k(\mathbf{x}_i, \mathbf{x}_j) = \mathrm{exp}\left( \dfrac{\| \mathbf{x}_i - \mathbf{x}_j \|^{2}}{2h^2} \right),
\end{equation}
where $h$ is a length scale parameter. 
Kernel PCA has proven a powerful nonlinear method, capable of overcoming the limitations of standard PCA without a significant increase in computational cost \cite{scholkopf1997kernel}. \review{In cases of very high-dimensional data, kernel-based methods rely highly on the efficiency of the chosen kernel and the corresponding distance metric.  To handle such cases, one can project high-dimensional data onto a lower-dimensional Riemannian manifold and define a valid kernel on that manifold \cite{dos2022grassmannian}. The dimensionality of projected data can be then further reduced via a standard DR method.}

\subsubsection{Isomap}
Isometric feature mapping (Isomap) \cite{tenenbaum2000global} is an extension of the multidimensional scaling (MDS) algorithm \cite{cox2008multidimensional}, which aims to identify a lower dimensional embedding in $\mathbb{R}^d$, $d \ll D$, that best preserves the geodesic distances between points in the ambient space. The goal is to minimize the objective function
\begin{equation}
\label{eq:isomap}
    \sum_{i,j,i \ne j} \big(d_{\mathcal{G}}(\mathbf{x}_i, \mathbf{x}_j) - d_\mathcal{E}(\mathbf{x}_i, \mathbf{x}_j)\big)^{2},
\end{equation}
where $d_{\mathcal{G}}$ represents the approximate geodesic distance and $d_\mathcal{E}$ is the Euclidean distance between the data points.

Isomaps begins by defining a neighborhood graph $\mathcal{G}$ over all data points. There are two primary variations of the Isomaps defined by how the graph is constructed. In the $\epsilon$-Isomap the graph is constructed by connecting points $i,j$ if $d(\mathbf{x}_i, \mathbf{x}_j) < \epsilon$, where $\epsilon$ is a fixed distance among vertices. Alternatively, the $k$-Isomap connects the $k$-nearest neighbors. Next, the shortest paths along the graph, the geodesic distances, between all pairs of nodes are computed. For each pair, this is typically done recursively using Floyd's algorithm \cite{tenenbaum2000global, floyd1962algorithm}. First the distances are initialized as $d_{\mathcal{G}}(\mathbf{x}_i, \mathbf{x}_j) = d(\mathbf{x}_i, \mathbf{x}_j)$ if $i,j$ are linked by an edge, or $d_{\mathcal{G}}(\mathbf{x}_i, \mathbf{x}_j) = \infty$ otherwise. Next, the matrix of shortest paths $\mathbf{D}_{\mathcal{G}}$ is computed recursively where, for $k=1,...,N$, each term is determined by
\begin{equation}
\label{eq:geod}
     d_{\mathcal{G}}(\mathbf{x}_i, \mathbf{x}_j) = \text{min}\{d_{\mathcal{G}}(\mathbf{x}_i, \mathbf{x}_j), d_{\mathcal{G}}(\mathbf{x}_i, \mathbf{x}_k) + d_{\mathcal{G}}(\mathbf{x}_k, \mathbf{x}_j)\}.
\end{equation}
Note that Floyd's algorithm requires $\mathcal{O}(N^3)$ operations. It is typically a good choice for dense graphs, but alternative methods such as Dijkstra's algorithm \cite{dijkstra1959note} may be preferable for sparse graphs. 


Lastly, the coordinates in the lower $d$-dimensional embedded space $(\mathbf{y}_1,...,\mathbf{y}_N)$ are calculated via the classical MDS algorithm \cite{cox2008multidimensional}, where first a double centering is applied to the shortest path matrix as
\begin{equation}
\label{eq:double-center}
    \mathbf{K} = -\frac{1}{2} \mathbf{H} \mathbf{D}_{\mathcal{G}} \mathbf{H},
\end{equation}
where $\mathbf{H} = \mathbf{I} - \frac{1}{N}\mathbf{J}_N$ is a centering matrix, $\mathbf{I}$ is the $N \times N$ identity matrix and $\mathbf{J}_N$ is a $N \times N$ matrix of all ones. Eigendecomposition of $\mathbf{K}$ is performed as
\begin{equation}
\label{eq:eigen-decomp}
    \mathbf{K} = \mathbf{W} \boldsymbol\Lambda \mathbf{W}^{\text{T}},
\end{equation}
and the $d$ largest eigenvalues $(\lambda_1, .., \lambda_d)$ of $\boldsymbol\Lambda$ and eigenvectors $(\mathbf{w}_1,..,\mathbf{w}_d)$ of $\mathbf{W}$ are retained. The dimension $d$ of the embedded space is determined through an adequate tolerance on the eigenvalues. Finally, the mapped data points onto the lower-dimensional space $\mathbf{Y} = \{ \mathbf{y}_1,..., \mathbf{y}_N\}$, where $\mathbf{y}_i \in \mathbb{R}^{d}$, are computed as follows
\begin{equation}
\label{eq:mapping}
    \mathbf{Y} = \mathbf{W}_d \boldsymbol\Lambda_d^{1/2},
\end{equation}
where $\mathbf{W}_d \in \mathbb{R}^{N \times d}$ and $\boldsymbol\Lambda \in \mathbb{R}^{d \times d}$.

\subsubsection{Diffusion maps}

Diffusion maps (DMaps) was introduced by Coifman and Lafon \cite{coifman2006diffusion} and is based on defining a Markov random walk on a graph over the data. Diffusion maps constructs a measure of proximity between data points, the so-called diffusion distance, by performing a random walk for a number of time steps. In the embedded space, the pairwise diffusion distances are optimally preserved. In contrast to other distance metrics such as the geodesic distance (employed by Isomap), the diffusion distance is considered more robust because it is based on integrating over all paths through the graph \cite{van2009dimensionality}.

As with other graph-based methods, DMaps begins by constructing a graph over the data. The edge weights in the graph are computed using a positive semi-definite kernel (see Section 2.1.1) where again the most widely-used kernel is the Gaussian kernel, given in Eq.~\eqref{eq:gaussian}. A random walk $W_N$ on the graph is considered and the transition probability matrix $\mathbf{P}$ is constructed as follows. First, the degree matrix is constructed as follows
\begin{equation}
    \text{D}_{ii} = \sum_{j=1}^{N} k(\mathbf{x}_i, \mathbf{x}_j),
\label{eq:diagonal}
\end{equation}
where $\text{D}_{ii}$ are the elements of a diagonal matrix $\mathbf{D} \in \mathbb{R}^{N \times N}$.
Next, the stationary probabilities of the random walk are defined as
\begin{equation}
    \pi_i = \frac{\text{D}_{ii}}{\sum\limits_{k=1}^{N} \text{D}_{kk}}.
\label{eq:stationary}
\end{equation}
Next, the kernel is normalized as
\begin{equation}
    \kappa_{ij} = \frac{k_{ij}}{\sqrt{\text{D}_{ii} \text{D}_{jj}}},
\label{eq:normalized}
\end{equation}
and the transition probability matrix $P_{ij}$ of the random walk is given by
\begin{equation}
    P_{ij} = \frac{\kappa_{ij}}{\sum\limits_{k=1}^{N} \text{D}_{ik}}.
\label{eq:trans_prob}
\end{equation}
It follows that the $t$-step transition probability matrix is given by $\mathbf{P}^t$, where $t$ now has the dual role of a time parameter (on the transition probability) and a scale parameter (on the graph). It has been shown, using spectral theory on the random walk \cite{lafon2006diffusion}, that the embedded space of points $\mathbf{Y}$, best retains the diffusion distances when it is formed by the $d$ non-trivial eigenvectors of $\mathbf{P}$. Performing the eigendecomposition of $\mathbf{P}$ and retaining the first $d$ eigenvalues $(\lambda_1, .., \lambda_d)$ of $\Lambda$ and eigenvectors $(\mathbf{w}_1,..,\mathbf{w}_d)$ yields a low-dimensional representation of points embedded in a Euclidean space given by
\begin{equation}
    \mathbf{Y} = \{\lambda_2^t \mathbf{w}_2, .., \lambda_{d+1}^t \mathbf{w}_{d+1} \}.
\label{eq:diff-coord}
\end{equation}
We note that because the graph is fully connected, the largest eigenvalue $\lambda_1$ is trivial, and thus its corresponding eigenvector $\mathbf{w}_1$ is discarded. Finally, it has been proven that the Euclidean distance in diffusion coordinates 
\begin{equation}
    D_t(\mathbf{x}_i, \mathbf{x}_j) =  \|\mathbf{Y}_i - \mathbf{Y}_j\|^2.
\end{equation}
approximates the diffusion distance on the graph, illustrating that the notion of proximity is well preserved in DMaps.

Two important variants of the DMaps that extend the method to increasingly complex and high-dimensional data sets (e.g. images and large matrices) are the vector diffusion maps \cite{singer2012vector}, which is derived using a heat kernel on vector fields, and the Grassmannian diffusion maps \cite{dos2022grassmannian, kontolati2021manifold, dos2021grassmannian2}, which embeds the structure of subspace data on the Grassmann manifold into a Euclidean space through an appropriate Grassmannian kernel.

\subsubsection{Locally linear embedding (LLE)}
Locally linear embedding (LLE) \cite{roweis2000nonlinear, polito2002grouping} is a sparse spectral technique that operates by implicitly fitting local hyperplanes to neighborhoods in the data. LLE then solves a sparse generalized eigenproblem to identify reduced coordinates that retain the local structure of the data \cite{van2009dimensionality}.  By preserving the local properties of the data points, non-convex manifolds can be also handled successfully. These local properties are specifically identified by expressing high-dimensional data points as a linear combination of their nearest neighbors such that, in the embedded space, the reconstruction weights in the linear combinations are retained.

The core assumption of LLE and related methods is that, for a sufficient number of data, each data point and its neighbors lie on or close to a locally linear patch of the manifold. As a first step, LLE identifies the nearest neighbors of each data point, by using for example the $k$-nearest neighbor algorithm or by choosing all points within a fixed radius. Each point in the dataset $\mathbf{x}_i$, is then expressed as the linear combination of its neighbors such that $\mathbf{x}_i \approx \tilde{\mathbf{x}}_i = \sum_{i=1}^{N} w_{ij} \mathbf{x}_j$. In effect, a hyperplane is fitted through the data point and its neighbors. The coefficients $w_{ij}$ are computed by minimizing the following quadratic cost function
\begin{equation}
    \epsilon(\mathbf{W}) = \sum_{i=1}^N \|\mathbf{x}_i - \sum_{j=1}^{N} w_{ij} \mathbf{x}_j \|^2 ,
\label{eq:weights-LLE}
\end{equation}
where $w_{ij}=0$ if $\mathbf{x}_j$ is not a neighbor of $\mathbf{x}_i$ and $\sum_{j=1}^{N} w_{ij} = 1, \hspace{5pt} \forall j$. This least-squares problem can be solved in closed form and the weights $\mathbf{W}$ are invariant to translation, rotation, and scaling, which results from the local linearity assumption. These properties allow linear mappings of the hyperplanes to a space of reduced dimensionality while preserving the reconstruction weights. The lower-dimensional data representation $\mathbf{Y}$ is then computed by minimizing the following cost function
\begin{equation}
    \phi(\mathbf{Y}) = \sum_{i=1}^N \|\mathbf{y}_i - \sum_{j=1}^{k} w_{ij} {\mathbf{y}_i}_j \|^2 \hspace{5pt} \text{subject to} \hspace{5pt} \| \mathbf{y}^{(k)} \|^2 = 1, \hspace{5pt} \forall k
\label{eq:cost-LLE}
\end{equation}
where $\mathbf{y}^{(k)}$ denotes the $k$-th column of matrix $\mathbf{Y}$. We note that the trivial solution $\mathbf{Y}=0$, is discarded through a constraint on the covariance of the columns of $\mathbf{Y}$, and translation invariance is imposed on the data (i.e., $\sum_{i}\mathbf{Y}_i= 0$). These two constraints ensure uniqueness of the solution. Roweis and Saul \cite{roweis2000nonlinear} have shown that 
\begin{equation}
    \phi(\mathbf{Y}) = \mathrm{tr}(\mathbf{Y}^{\text{T}} \mathbf{M} \mathbf{Y}),
\label{eq:res-LLE}
\end{equation}
where $\mathbf{M} = (\mathbf{I} - \mathbf{W})^{\text{T}} (\mathbf{I} - \mathbf{W})$. Therefore, $\phi(\mathbf{Y})$ is minimized by solving for the $d$ eigenvectors associated with the $d$ smallest eigenvalues of $\mathbf{M}$ (excluding the first one with a zero eigenvalue). Similarly to the other spectral methods presented, the reduced dimensionality $d$ can be chosen based on an adequate tolerance on the eigenvalues. 

Some related methods build on the LLE method including the Hessian LLE \cite{donoho2003hessian} that modifies the LLE to use Hessian matrices, and the Local Tangent Space Alignment (LTSA) \cite{zhang2004principal} that operates by aligning the local tangent spaces to unfold the manifold on which the data lie.

\subsubsection{Laplacian Eigenmaps}
Similarly to LLE, Laplacian Eigenmaps \cite{belkin2003laplacian}  aims to preserve local properties of the manifold. The goal is to identify a lower-dimensional embedding by minimizing distances between each data point and its $k$-nearest neighbors. First, a neighboring graph $\mathcal{G}$ is constructed where, for all points connected by an edge, the weight is calculated using the Gaussian kernel in Eq.~\eqref{eq:gaussian}. From these weights, a sparse adjacency matrix $\mathbf{W}$ is constructed. To identify the lower-dimensional representation $\mathbf{Y}$, the following cost function is minimized
\begin{equation}
    \phi(\mathbf{Y}) = \sum_{ij} \| \mathbf{y}_i - \mathbf{y}_j \|^2 w_{ij},
\label{eq:cost1-LE}
\end{equation}
which enforces points that lie close in the original space to be similarly close in the embedded space. Notice the similarity to Eq.\ \eqref{eq:cost-LLE} where the main difference is that weights are applied to the edges of the graph, as opposed to the weights applying directly to the neighboring points.

The above minimization problem can be reposed as an eigenproblem by first computing the degree matrix $\mathbf{M}$ (where $m_{ii} = \sum_{j} w_{ij}$) and the graph Laplacian $\mathbf{L}$ (where $\mathbf{L} = \mathbf{M} - \mathbf{W}$), which approximates the Laplace-Beltrami operator on the manifold. The cost function can then be written as $\phi(\mathbf{Y}) = 2\mathbf{Y}^{\text{T}} \mathbf{L} \mathbf{Y}$. Hence, minimizing $\phi(\mathbf{Y})$ is equivalent to minimizing $\mathbf{Y}^{\text{T}} \mathbf{L} \mathbf{Y}$ subject to $\mathbf{Y}^{\text{T}} \mathbf{M} \mathbf{Y} = \mathbf{I}_N$ where $\mathbf{I}_N$ is the $N \times N$ identity matrix. Therefore the lower-dimensional representation of the data can be computed by solving the following eigenvalue problem
\begin{equation}
    \mathbf{L} \mathbf{v} = \lambda \mathbf{M} \mathbf{v},
\label{eq:eigen-LE}
\end{equation}
where the lower-dimensional coordinates $\mathbf{v}_0, \dots, \mathbf{v}_{d-1}$ are the solutions to Eq.\ \eqref{eq:eigen-LE} corresponding to the $d$ smallest nonzero eigenvalues such that a point $\mathbf{y}_i$ in the lower dimensional space is given by the coordinates $\mathbf{y}_i = \{\mathbf{v}_0(i), \dots, \mathbf{v}_{d-1}(i)\}$. The dimension of the reduced-space is computed similarly to the spectral methods presented above.

\subsection{Blind source separation methods}
Blind source separation (BSS) is a class of unsupervised learning methods used in many applications including audio signal processing, image processing, and electroencephalography (EEG), based on the assumption that a complex signal can be decomposed into a linear combination of a priori unknown mutually independent sources \cite{pal2013blind}. A classic example of source separation is the Cocktail Party problem, which involves identifying a particular sound of interest, often a speech signal, in a complex auditory setting with many additional signals (i.e. several concurrent conversations) \cite{qian2018past}. The problem of BSS boils down to identifying a representation in which the total source is expressed as linear combination of component signals that are statistically independent, or as independent as possible.

\subsubsection{Independent Component Analysis (ICA)}

Given $N$ realizations of a $D$-dimensional random vector $\mathbf{X}=\{x_1(t),x_2(t),...,x_N(t)\}^{\text{T}}$, where $t$ represents time or another sample index, independent component analysis (ICA) \cite{hyvarinen2000independent} seeks to identify the linear transformation $\mathbf{W}$ such that
\begin{equation}
\begin{pmatrix}
s_1(t)\\
s_2(t)\\
\vdots \\
s_N(t)
\end{pmatrix}
= \mathbf{W}
\begin{pmatrix}
x_1(t)\\
x_2(t)\\
\vdots \\
x_N(t)
\end{pmatrix},
\label{eq:ICA}
\end{equation}
where $\mathbf{S}=\{s_1(t),s_2(t),...,s_N(t)\}^{\text{T}}$ is a vector of maximally independent components. Estimation of $\mathbf{W}$ is based on the principle of maximum non-Gaussianity. According to the central limit theorem (CLT), the sum of non-Gaussian random variables is closer to a Gaussian than the original ones. Therefore, signal mixtures tend to have Gaussian PDFs, while source signals tend to have non-Gaussian PDFs. The vectors of matrix $\mathbf{W}$, also known as the \textit{unmixing} matrix, are calculated adaptively by maximizing the non-Gaussianity in $\mathbf{S}$. A common measure of non-Gaussianity is the kurtosis \cite{hyvarinen2000independent} of a random variable $X$, a fourth-order statistic defined as
\begin{equation}
\text{kurt}(X) = \mathbb{E}\{X^4\} - 3 [\mathbb{E}\{X^2\}]^2,
\label{eq:kurt}
\end{equation}
where $\text{kurt}(X)=0$ if $X$ is Gaussian. Maximizing non-Gaussianity can be achieved by applying gradient-based algorithms that search for a matrix $\mathbf{W}$ such that the projection $\mathbf{W}\mathbf{X}$ has maximum kurtosis. A fixed-point algorithm known as FastICA \cite{hyvarinen1997fast, bingham2000fast, hyvarinen1999fast} has been proposed to accelerate convergence. The algorithm starts from a set of vectors $\mathbf{W}$, computes the direction in which the absolute value of the kurtosis is growing most rapidly, based on the available samples of mixture vector $\mathbf{X}$, and then moves $\mathbf{W}$ in that direction \cite{hyvarinen2000independent}. After optimization, the estimated sources can be obtained by transforming the data with the estimated unmixing matrix to obtain the $d$-dimensional reduced representation.

\subsubsection{Non-Negative Matrix Factorization (NMF)}

Non-negative matrix factorization or NMF, is another BSS algorithms also often used for feature extraction \cite{paatero1994positive}. NMF uses concepts from multi-variate analysis and linear algebra and  employs the non-negativity constraint which is natural in many physical problems, for example, in astronomy, chemometrics, computer vision, and audio signal processing. 

Consider $N$ observations of the random vector $\mathbf{x} \in \mathbb{R}^D$ and let the data matrix be $\mathbf{X} = [\mathbf{x}_1, \mathbf{x}_2, .., \mathbf{x}_N]^{\text{T}} \in \mathbb{R}_{\ge 0}^{N \times D}$. NMF decomposes $\mathbf{X}$ into a product of a non-negative basis matrix $\mathbf{G} \in \mathbb{R}_{\ge 0}^{N \times d}$ and a non-negative coefficient matrix $\mathbf{F} \in \mathbb{R}_{\ge 0}^{d \times M}$, such that $\mathbf{X} \approx \mathbf{G} \mathbf{F}$ (or equivalently $\mathbf{x}_j \approx \sum_{i=1}^{d}g_i F_{ij}$)
\cite{wang2012nonnegative}, thus decomposing each data point into a linear combination of the basis vectors. The NMF representation is determined by minimizing the Frobenius norm between $\mathbf{X}$ and $\mathbf{G}\mathbf{F}$ as
\begin{equation}
\{\mathbf{G},\mathbf{F}\} = \underset {\mathbf{G},\mathbf{F}}{\arg\min} \| \mathbf{X} - \mathbf{G} \mathbf{F} \|_{\text{F}},
\label{eq:NMF}
\end{equation}
where matrices $\mathbf{G}$ and $\mathbf{F}$ are required to be of previously selected rank $d$. The reduced dimension/rank $d$ can be chosen such that the smallest possible Frobenius norm is computed. Lee and Seung \cite{lee1999learning, fevotte2011algorithms} also proposed an alternative cost function, using the extension of the Kullback-Leibler (KL) divergence to positive matrices. NMF has the ability to generate factor matrices of significantly reduced dimensionality compared to the original data matrix. The factorization of NMF is not unique and more control over the non-uniqueness can be obtained with sparsity constraints \cite{eggert2004sparse}.

\subsection{Non-convex methods}

In this Section, we present three techniques that construct a lower-dimensional representation through optimizing a non-convex function, a method based on the minimization of data distributions used for visualization and two neural network-based methods.

\subsubsection{t-distributed Stochastic Neighbor Embedding (t-SNE)}

The t-distributed stochastic neighbor embedding (t-SNE) \cite{van2008visualizing} is a nonlinear DR technique used primarily for visualization of high-dimensional data in 2- or 3-dimensional scatter plots. The main advantage of t-SNE, compared to other embedding approaches, is its ability to retain both the local and the global structure of the data in a single map. The goal of t-SNE is to minimize the divergence between the two distributions that measure pairwise similarities of points in the original and the reduced space respectively. 

Consider a set of high-dimensional input samples $\mathbf{X} = \{\mathbf{x}_{1},..,\mathbf{x}_{N}\}^{\text{T}}$ where $\mathbf{x}_i \in \mathbb{R}^D$ and a function $d(\mathbf{x}_i,\mathbf{x}_j)$ that computes the pairwise distances between data points (e.g., the Euclidean distance $d(\mathbf{x}_i,\mathbf{x}_j)=\|\mathbf{x}_i-\mathbf{x}_j\|$). The following conditional probabilities measure the pairwise similarity between points $\mathbf{x}_i$,$\mathbf{x}_j$ \cite{van2014accelerating}
\begin{subequations}
\begin{align}
    p_{j|i} &= \frac{\exp{(-d(\mathbf{x}_i,\mathbf{x}_j)^2/2\sigma_i^2)}}{\sum_{k \ne i} {\exp{(-d(\mathbf{x}_i,\mathbf{x}_k)^2/2\sigma_i^2)}}}, \hspace{15pt} p_{i|i}=0 \\
    p_{ij} &= \frac{p_{j|i} + p_{i|j}}{2N},
\end{align}
\label{eq:t-sne}
\end{subequations}
where $\sigma_i$ is the bandwidth of the Gaussian kernel. Similarities between points in the $d$-dimensional embedding $\mathbf{y}_i, \mathbf{y}_j$ are computed using a normalized Student-t kernel with a single degree of freedom as
\begin{equation}
    q_{ij} = \frac{(1 + \|\mathbf{y}_i - \mathbf{y}_j \|^2)^{-1}}{\sum_
    {k \ne l} (1+ \|\mathbf{y}_k - \mathbf{y}_l \|^2)^{-1}}, \hspace{15pt} q_{ii}=0.
\label{eq:t-sne-student}
\end{equation}
The heavy-tailed distribution allows dissimilar points in the original space to be modeled with points that are far apart in the embedding. The embedded coordinates, $\mathbf{Y} = \{\mathbf{y}_{1},..,\mathbf{y}_{N}\}$ where $\mathbf{y}_i \in \mathbb{R}^d$ and $d \ll D$, are solved by minimizing the KL divergence (also known as relative entropy) between the joint distributions $P$ and $Q$ given by
\begin{equation}
    C(\mathbf{Y}) = \text{KL}(P\|Q) = \sum_{i \ne j} \log \frac{p_{ij}}{q_{ij}},
\label{eq:t-sne-KL}
\end{equation}
This objective function is generally non-convex and is typically minimized with gradient descent-based algorithms, where the gradient can be determined by
\begin{equation}
    \frac{\partial C}{\partial \mathbf{y}_i} = 4 \sum_{j \ne i}(p_{ij}- q_{ij}) q_{ij} Z(\mathbf{y}_i - \mathbf{y}_j),
\label{eq:t-sne-KL-grad}
\end{equation}
where $Z = \sum_{k \ne l} (1+ \|\mathbf{y}_k - \mathbf{y}_l \|^2)^{-1}$ is a normalization term. The main limitation of t-SNE is its computational complexity scales quadratically with the number of input samples $N$, i.e., $\mathcal{O}(N^2)$. For this reason, tree-based approaches have been proposed to accelerate convergence by constructing approximate similarities which can lead to a complexity of $\mathcal{O}(N\log{N})$ \cite{van2014accelerating}.


\subsubsection{Autoencoders}

Artificial neural networks (ANNs) have been established as universal function approximators and have been extensively used for supervised learning tasks. In fact, it has been proven that a single hidden layer ANN, can approximate accurately any nonlinear continuous function \cite{chen1995universal, rossi2005functional}. Mathematically, a single hidden layer ANN has the following representation
\cite{bank2020autoencoders}
\begin{equation}
    \mathbf{h} = \sigma(\mathbf{W}\mathbf{x} + \mathbf{b}),
\label{eq:ann}
\end{equation}
where $\mathbf{x}$ represents the input quantity, $\mathbf{W}$ and $\mathbf{b}$ are the trainable weights and biases respectively and $\sigma(\cdot)$ is a nonlinear activation function. Any ANN with more than one hidden layer is known as a deep neural network (DNN). 

A class of DNNs with bottleneck architecture, known as \textit{autoencoders} first introduced by Rumelhart et al.\  \cite{rumelhart1985learning}, can be used to learn efficient encodings of unlabeled data, an unsupervised learning task. Their main objective is to learn an optimally informative and compact representation of the data that allows both the reduction and reconstruction of high-dimensional datasets. Assuming an autoencoder network, the objective is to learn the encoder and decoder mappings, $Q: \mathbb{R}^D \rightarrow \mathbb{R}^d$ and $G: \mathbb{R}^d \rightarrow \mathbb{R}^D$ respectively, satisfying the following \cite{baldi2012autoencoders}
\begin{equation}
    \underset {Q,G}{\arg\min} \mathbb{E}[\Delta(\mathbf{x}, G \circ Q(\mathbf{x})],
\label{eq:autoencoder}
\end{equation}
where $\mathbb{E}$ is the expectation over the distribution of random variable $x$ and $\Delta$ is the reconstruction loss function which represents the ``distance'' between the input and the output of the decoder. This is usually defined using a mean-squared error.

In the case where $Q,G$ are linear operators, the encoder produces the same latent representation as PCA (see Section \ref{PCA}). However, the nonlinear activation functions $\sigma(\cdot)$ allow for a superior reconstruction when compared to PCA. The number of neurons in the last hidden layer of the encoder $Q$ represents the dimension $d$ and the reduced solution is obtained by directly applying the trained encoder to the input data. The reduced dimension $d$ can be chosen, such that the quantity in Eq.~(\ref{eq:autoencoder}) is minimized. Altering the network architecture and objective results in different methods that often allow the extraction of more complex latent codings. Furthermore, bias-variance tradeoff can be addressed by suing regularized and sparse autoencoders which enforce sparsity on the hidden activations \cite{bank2020autoencoders}.

\subsubsection{Wasserstein Autoencoders (WAE)}

Standard autoencoders are trained by the mean-squared error (MSE) loss function which aims to match inputs to the resulting decoder outputs. By interpolation on the latent space, one can generate new samples which can be reconstructed via the decoder. However, the MSE loss function does not take into account how realistic these samples are. To tackle this issue and achieve smooth interpolation, variational autoencoders (VAEs) have been proposed \cite{kingma2013auto, kingma2019introduction}. A VAE provides a probabilistic representation of the observations and enforces the latent representation to resemble a prior, usually a multivariate standard normal. This is achieved via the minimization of the reconstruction loss \textit{and} the KL divergence which, as discussed previously, represents a measure of dissimilarity between two distributions. 

Wasserstein autoencoders (WAE) \cite{tolstikhin2017wasserstein, rubenstein2018latent} minimize the Wasserstein distance, a special case of the optimal transport (OT) distance between two probability distributions $P_X$ and $P_G$ defined as 
\cite{baldi2012autoencoders}
\begin{equation}
    W_c(P_X, P_G) = \inff{\Gamma \in P(X \sim P_X, Y \sim P_G)} \mathbb{E}_{(X,Y)\sim \Gamma}[c(X,Y)]
\label{eq:wae_dist}
\end{equation}
where $c(x,y)$ denotes some cost function. When $c(x,y)=d^p(x,y)$, where $d(x,y)$ is a distance metric, then the $p$-th root of $W_c$ is called the $p$-Wasserstein distance. In the special case where $p=1$ and $c(x,y)=d(x,y)$ we get the 1-Wasserstein distance also known as the Earth mover's distance \cite{rubner2000earth}. If we assume that $Q$ is the encoder and $G$ the decoder, a WAE is trained by the following objective function
\begin{equation}
    D_{\text{WAE}}(P_X,P_G)= \inff{Q(Z|X)_{\in \mathscr{Q}}} \mathbb{E}_{P_X} \mathbb{E}_{Q(Z|X)} [c(X,G(Z))] + \lambda D_Z(Q_Z,P_Z),
\label{eq:wae_obj}
\end{equation}
where $z$ represents the latent variable. The left part of the RHS of Eq.~\eqref{eq:wae_obj}, represents the reconstruction loss whereas the right part, penalizes the distance between the latent space distribution to the
prior distribution. As pointed out by Tolstikhin et al.\ \cite{tolstikhin2017wasserstein}, WAEs share many properties with VAEs but generate samples of better quality. Similarly to autoencoders, the reduced dimension $d$ in WAEs, is controlled by the number of neurons in the last hidden layer of the encoder network.

\section{High-dimensional uncertainty quantification framework}
\label{S:approach}

\begin{figure}[ht!]
\begin{center}
\includegraphics[width=0.75\textwidth]{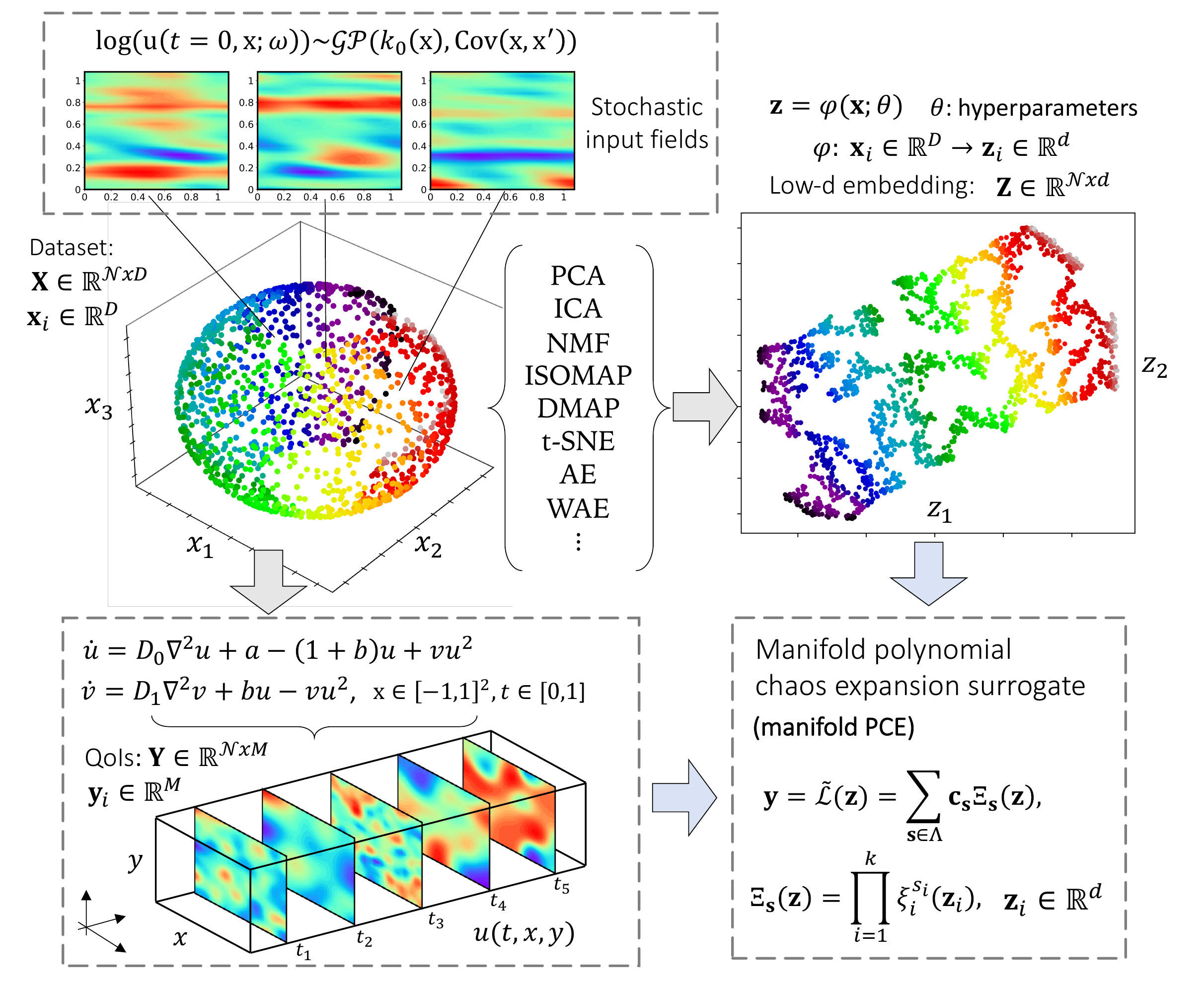}
\caption{A schematic illustrating the m-PCE approach. The construction of a PCE surrogate whose inputs are projected on a lower-dimensional manifold, enables UQ in physics-based models associated with high-dimensional stochastic input fields.}
\label{fig:schematic}
\end{center}
\end{figure}

Consider an analytical or computational model $\mathcal{M}$ which simulates a physical process and represents a mapping between input random variables, and output QoIs. We generate $N$ input realizations $\mathbf{X} = \{\mathbf{x}_{1},..,\mathbf{x}_{N}\}$ and compute associated QoIs $\mathbf{Y} = \{\mathbf{y}_{1},..,\mathbf{y}_{N}\}$, with model $\mathcal{M}: {\mathbf{x}_{i} \in \mathbb{R}^{D}} \rightarrow \mathbf{y}_{i} \in {\mathbb{R}^{M}}$. We assume that the dimensionality, $D$, of the input random variables is high, e.g. on the order of $\mathcal{O}(10^{2-4})$. The high-dimensional inputs can represent random fields and processes, for example, such as spatially or temporally varying coefficients. Our first objective is to identify a lower-dimensional representation of the input data points whose mapping to the output QoIs will be approximated via the construction of a PCE surrogate model. A graphical summary of the m-PCE approach is shown in Figure \ref{fig:schematic}. 

\subsection{Dimension reduction of high-dimensional inputs}

The problem of dimensionality reduction is defined as follows. Given a high-dimensional dataset $\mathbf{X} = \{\mathbf{x}_{1},..,\mathbf{x}_{N}\}$ where $\mathbf{x}_i \in \mathbb{R}^D$, assume the intrinsic dimensionality of the dataset $d$ ($d \approx \mathcal{O}(10^{0-1})$), where $d<D$ and often $d \ll D$. The existence of an intrinsic dimensionality assumes that data points lie on or near a manifold of dimensionality $d$ that is embedded in the $D$-dimensional space. The structure of the $d$-dimensional manifold is unknown and it may even consist of a number of disconnected submanifolds \cite{van2009dimensionality}. The goal of DR is to identify a low-dimensional representation of the original dataset $\mathbf{Z} = \{\mathbf{z}_{1},..,\mathbf{z}_{N}\}$, $\mathbf{z}_i \in \mathbb{R}^d$, while retaining its meaningful properties. We denote this mapping as $\mathbf{z}=\phi(\mathbf{x};\theta)$, where $\phi: \mathbf{x}_i \in \mathbb{R}^D \rightarrow \mathbf{z}_i \in \mathbb{R}^d$ and $\theta \in \mathbb{R}^n$ represents the $n$-dimensional vector of parameters associated with the reduction method. The resulting dimension $d$ should ideally be close to the intrinsic dimension, however since this number is unknown a-priori, the problem is ill-posed and requires certain assumptions to be solved. The process of DR involves some transformation operation from the original to the reduced feature space \cite{sorzano2014survey}. Although certain methods allow an inverse transformation $\phi^{-1}: \mathbf{z}_i \in \mathbb{R}^d \rightarrow \mathbf{x}_i \in \mathbb{R}^D $ to map data back to the original space, for others this inverse mapping may not exist. Mathematically, to obtain the low-dimensional representation, an error measure $L(\mathbf{x};\theta)$ needs to be defined and minimized to estimate the optimal parameters $\hat{\theta} = \underset {\theta}{\arg\min} L(\mathbf{x};\theta)$. For example, when the inverse transformation $\phi^{-1}$ is available and the objective is data reduction without information loss, this function can be represented by the mean-squared reconstruction error \cite{lataniotis2020extending}. 

Several methods studied in this work are considered convex, since the objective function does not contain local minima. The low-dimensional data representation is obtained by minimizing the convex objective function by means of an eigendecomposition. These include the PCA, kernel-PCA, ISOMAP, and DMAP. Sparse spectral techniques such as LLE and Laplacian eigenmaps focus on solving a sparse generalized eigenproblem which translates in retaining the local structure of the data. Non-convex techniques include multi-layer neural networks such as autoencoders. Blind source separation techniques such as ICA or NMF treat the high-dimensional dataset as a mixture of signals and aims to identify its individual sources. 

The quality of the compressed data can be assessed by evaluating to what extent the local structure of the data is preserved. This evaluation can be performed by measuring the trustworthiness and the continuity of the low-dimensional embeddings. The reader is referred to \cite{van2009dimensionality} for more information.


\subsection{Uncertainty propagation}
Having defined a framework for identifying a low-dimensional representation of input random variables, we consider an unknown, closed form function $\mathcal{L}$, which allows us to map the reduced uncertain inputs to the QoIs as $\mathbf{y} = \mathcal{L}(\mathbf{z}) = \mathcal{L}(\phi(\mathbf{x};\theta))$ where $\mathcal{L}:\mathbb{R}^d \rightarrow \mathbb{R}^M$. Let us assume that the $d$-dimensional reduced input random vector is formalized using an appropriate probability distribution $\mathbf{z} \sim p(\mathbf{z})$. The main goal of uncertainty propagation is to characterize the statistics of $\mathcal{L}(\mathbf{z})$. The first two moments, for example, are given by
\begin{subequations}
\begin{align}
    \mu_{\mathcal{L}} &= \int \mathcal{L}(\mathbf{z})p(\mathbf{z})d\mathbf{z} \\
    \sigma^2_{\mathcal{L}} &= \int (\mathcal{L}(\mathbf{z}) - \mu_{\mathcal{L}})^2 p(\mathbf{z})d\mathbf{z},
\end{align}
\label{eq:up-moments}
\end{subequations}
The above integrals can be approximated via standard Monte Carlo simulation. By the law of large numbers, the approximate mean $\hat{\mu}_{\mathcal{L}}$ and variance $\hat{\sigma}^2_{\mathcal{L}}$, converge to the true values as the samples size tends to infinity. For computationally demanding high-fidelity models, which is often the case in physics-based problems, a large number of evaluations is infeasible. We note, that these are essentially evaluations of the original model $\mathcal{M}$ as $\mathbf{y}=\mathcal{M}(\mathbf{x})=\mathcal{L}(\phi(\mathbf{x};\theta))$. Therefore, an approximate surrogate model $\widetilde{\mathcal{L}}$ is constructed using the set of reduced samples $\mathbf{z}$ and the corresponding model solutions $\mathbf{y}$ as training data. Once the surrogate model is appropriately trained it can be employed to inexpensively generate a very large number of approximate model evaluations in two steps: (1) the projection function $\phi: \mathbf{x}_i \in \mathbb{R}^D \rightarrow \mathbf{z}_i \in \mathbb{R}^d$ is used to encode new realizations of the high-dimensional inputs to a latent space and (2) the surrogate model $\widetilde{\mathcal{L}}: \mathbf{z}_i \in \mathbb{R}^d \rightarrow \mathbf{y}_i \in \mathbb{R}^M$ is used to map reduced data to the QoIs. 

\subsection{Surrogate modeling via polynomial chaos expansion}

To construct a surrogate model $\widetilde{\mathcal{L}}$ that emulates the mapping $\mathbf{y} = \mathcal{L}(\mathbf{z})$ from the latent space to the output space i.e.,\ $\mathcal{L}:\mathbb{R}^d \rightarrow \mathbb{R}^M$ we use non-intrusive PCE, an efficient and flexible surrogate modeling method. For the purposes of the method, the reduced input random variable $\mathbf{z}$, is considered to be defined on the probability space $\left(\Omega, \Sigma, P\right)$ and characterized by the joint probability density function (PDF) $\varrho_{\mathbf{Z}}: I \rightarrow \mathbb{R}_{\geq 0}$, where $I \subseteq \mathbb{R}^d$ is the image space, $\Omega$ the sample space, $\Sigma$ the set of events, and $P$ the probability measure. 

In the following, we consider for simplicity a single model output, such that $Y\left(\omega\right) = \mathcal{L}\left(\mathbf{z}\left(\omega\right)\right) \in \mathbb{R}$, $\omega \in \Omega$. Under this assumption, the PCE is a spectral approximation of the form
\begin{equation}
\label{eq:spectral_approx}
\mathcal{L}(\mathbf{z}) \approx \widetilde{\mathcal{L}}(\mathbf{z}) = \sum_{s=1}^S c_s \Xi_s(\mathbf{z}),
\end{equation}
where $c_s$ are scalar coefficients and $\Xi_s$ are multivariate polynomials that are orthonormal with respect to the joint PDF $\varrho_{\mathbf{Z}}$, such that
\begin{equation}
\label{eq:orthNd}
\mathbb{E}\left[\Xi_s \Xi_t\right] = \int_{S}  \Xi_s\left(\mathbf{z}\right) \Xi_t\left(\mathbf{z}\right) \varrho_{\mathbf{z}}\left(\mathbf{z}\right) \mathrm{d}\mathbf{z} =  \delta_{st},
\end{equation}
where $\delta_{st}$ denotes the Kronecker delta.
Depending on the PDF $\varrho_{\mathbf{z}}$, the orthonormal polynomials can be chosen according to the Wiener-Askey scheme \cite{xiu2002wiener} or be numerically constructed \cite{soize2004physical}.
The joint PDF of independent random variables $z_1, \dots, z_k$ and the multivariate orthogonal polynomials are given by
\begin{equation}
\label{eq:joint_pdf}
\varrho_{\mathbf{z}} \left(\mathbf{z}\right) = \prod_{i=1}^d \varrho_{z_i}\left(z_i\right), \hspace{10pt} \Xi_s(\mathbf{z}) \equiv \Xi_\mathbf{s}(\mathbf{z}) = \prod_{i=1}^d \xi_i^{s_i} (z_i),
\end{equation}
where $\varrho_{z_i}$ is the marginal PDF of random variable $z_i$ and $\xi_i^{s_i}$ are univariate polynomials of degree $s_i \in \mathbb{Z}_{\geq 0}$ and orthonormal with respect to the univariate PDF $\varrho_{z_i}$, such that
\begin{equation}
\label{eq:orth1d}
\mathbb{E}\left[\xi_i^{s_i}\xi_i^{t_i}\right] = \int_{z_i} \xi_i^{s_i}(z_i) \xi_i^{t_i}(z_i) \varrho_{z_i}(z_i) \, \mathrm{d}z_i = \delta_{s_i t_i}.
\end{equation}

The multi-index $\mathbf{s} = \left(s_1, \dots, s_k\right)$ is equivalent to the multivariate polynomial degree and uniquely associated to the single index $s$ employed in Eq.~\eqref{eq:spectral_approx}, which can now be written in the equivalent form
\begin{equation}
\label{eq:spectral_approx_multi_index}
\mathcal{L}(\mathbf{z}) \approx \widetilde{\mathcal{L}}(\mathbf{z}) = \sum_{\mathbf{s} \in \Lambda} c_{\mathbf{s}} \Xi_{\mathbf{s}}(\mathbf{z}),
\end{equation}
where $\Lambda$ is a multi-index set with cardinality $\#\Lambda = S$.
The choice of the multi-index set $\Lambda$ plays a central role in the construction of the PCE, as it defines which polynomials and corresponding coefficients form the PCE.
The most common choice, as well as the one employed in this work, is that of a total-degree multi-index set, such that $\Lambda$ includes all multi-indices that satisfy $\left\| \mathbf{s} \right\|_1 \leq s_{\max}$, $s_{\max} \in \mathbb{Z}_{\geq 0}$. In that case, the size of the PCE basis is $S = \frac{\left(s_{\max} + k\right)!}{s_{\max}! k!}$, i.e., it scales polynomially with the input dimension $k$ and the maximum degree $s_{\max}$.

Once the multi-index set $\Lambda$ is fixed, the only thing remaining to complete the PCE is to compute the coefficients, which is achieved here by means of regression \cite{blatman2011adaptive, loukrezis2020robust, hampton2018basis, he2020adaptive, tsilifis2019compressive}. In particular, the PCE coefficients are obtained by solving the penalized least squares problem  
\begin{align}
\label{eq:regression}
\underset {\mathbf{c} \in \mathbb{R}^{\#\Lambda}}{\arg\min} \left\{\frac{1}{N}\sum_{i=1}^{N} \left( \mathcal{L}(\mathbf{z}_i) - \sum_{\mathbf{s} \in \Lambda} c_{\mathbf{s}} \Xi_{\mathbf{s}}\left(\mathbf{z}_i\right) \right)^2 + \lambda J\left(\mathbf{c}\right)\right\},
\end{align}
where $\lambda \in \mathbb{R}$ is a penalty factor, $J\left(\mathbf{c}\right)$ is a penalty function acting on the vector of PCE coefficients $\mathbf{c} \in \mathbb{R}^{\#\Lambda}$, and $\mathbf{Z} = \left\{\mathbf{z}_i\right\}_{i=1}^{N}$ is an experimental design (ED) of random variable realizations with corresponding model outputs $\mathbf{Y} = \left\{\mathbf{y}_i\right\}_{i=1}^{N}$. Common choices for the penalty function $J(\mathbf{c})$ are the $\ell_1$ and $\ell_2$ norms, in which cases problem~\eqref{eq:regression} is referred to as LASSO (least absolute shrinkage and selection operator) and ridge regression, respectively. Removing the penalty term results in an ordinary least squares (OLS) regression problem.

\subsubsection{PCE on latent spaces}

\review{The theoretical challenges in applying the PCE to the reduced dimensional vector $\mathbf{z}$ are that its probability measure is not generally known exactly, it may take a non-standard form, and it may include dependencies between variables. Several techniques have been developed to address these challenges. For non-standard measures, orthogonal polynomials can be constructed by Stiltjes or Gram-Schmidt orthogonalization \cite{soize2004physical, wan2006beyond}. Furthermore, for non-standard or unknown measures the polynomials can be constructed to be orthogonal with respect to statistical moments rather than the joint pdf, using an approached termed arbitrary PCE (aPCE) \cite{oladyshkin2012data}. Finally, dependencies can either be ignored altogether or they can be explicitly modeled through copulas and a corresponding Rosenblatt transformation to obtain independent uniform random variables \cite{torre2019data}. }  

\review{However, for practical purposes the reduced input variables can be approximated as either uniformly or normally distributed and independent. Although this violates certain theoretical considerations, it does not lead to significant differences in the model accuracy as illustrated in the examples below. Furthermore, it has been shown in the literature that ignoring the input dependencies results in good surrogate model performance, since complex nonlinear input transformations, e.g., via the Rosenblatt transform, are avoided \cite{rosenblatt1952remarks,torre2019data, lataniotis2020extending}. Herein, we approximate the reduced input variables as independent and uniformly distributed over the support of the reduced data set. Again, we emphasize that this is done for simplicity and because it is demonstrated to give sufficiently accurate results. If these assumptions become problematic,  either due to theoretical reasons or due to a lack of predictive accuracy, one can always resort to the methods discussed above to improve the PCE approximation. There are no constraints against using these methods within the m-PCE framework.}

\review{Furthermore, in the proposed framework QoIs are vector-valued, usually representing fields that evolve in space and time, and the PCE approximation described in the previous Section is applied element-wise. The way the surrogate is formulated is that a basis is constructed, and the coefficients are computed for each output element (location in the field). The correlation between output grids is implicitly taken into account, as the learned sets of PCE coefficients are highly correlated. The m-PCE approach can be extended to perform dimension reduction in both input and output fields which results in the construction of very few PCE surrogates for output components that do not have the same spatial dependence \cite{kontolati2022influence}.}

\subsubsection{Error estimation of surrogate predictions}
\label{error-metrics}

To assess the predictive accuracy of the PCE surrogate, we employ two metrics. The first scalar metric is the relative $L_2$ error given by
\begin{equation}
\label{eq:L2-error}
    L_2(\mathbf{y}_{\text{pred}}, \mathbf{y}_{\text{ref}}) = \frac{\|\mathbf{y}_{\text{pred}}- {\mathbf{y}_{\text{ref}}\|}_{2}}{{\|\mathbf{y}_{\text{ref}}\|}_{2}},
\end{equation}
where ${\|\cdot\|}_2$ denotes the standard Euclidean norm and $\mathbf{y}_{\text{pred}}, \mathbf{y}_{\text{ref}}$ are the prediction and reference responses respectively.
As a second scalar metric we introduce is the $R^2$ score, also known as the coefficient of determination, defined as
\begin{equation}
\label{eq:R2-score}
    R^2 = 1 - \frac{\mathlarger{\sum}_{i=1}^{w} \Big(\mathbf{y}_{\text{pred}, i} - \mathbf{y}_{\text{ref}, i}\Big)^2}{\mathlarger{\sum}_{i=1}^{w} \Big(\mathbf{y}_{\text{ref}, i} - \overline{\mathbf{y}}_{\text{ref}}\Big)^2},
\end{equation}
where $w$ is the total number of mesh points of the QoI and $\overline{\mathbf{y}}_{\text{ref}}$ is the mean reference response. The above metrics are computed for a holdout test dataset of transformed inputs and model outputs $\{{\mathbf{z}^*_{i} \in \mathbb{R}^{d}}, \mathbf{y}^*_{i} \in {\mathbb{R}^{M}}\}_{i=1}^{N^*}$ where $N^*$ is the total number of test realizations.

\subsection{Hyperparameter optimization}
Hyperparameter optimization refers to the problem of optimizing a loss function over a graph-structured configuration space \cite{bergstra2011algorithms}. The mapping of high-dimensional inputs to a latent space, $\mathbf{z}=\phi(\mathbf{x};\theta)$ is associated with $n$ hyperparameters. The domain of the $\hat{n}$-th hyperparameter is denoted as $\Theta_{\hat{n}}$ and the overall configuration space as $\boldsymbol\Theta= \Theta_1 \times \Theta_2 \times \cdots \times \Theta_n$ thus the vector of hyperparameters is denoted as $\theta \in \boldsymbol\Theta$. In a more naive approach one resorts to extensive experimentation or grid search to obtain a suitable model which can be extremely time-consuming as it suffers from the curse of dimensionality. Hyperparameter optimization algorithms optimize variables which can be real (e.g., learning rate) or integer-valued (e.g., number of layers in an ANN), typically with bounded domains, binary (e.g., early stopping or not), or categorical (e.g., choice of the optimizer) \cite{feurer2019hyperparameter}. Given an infinite-sample dataset $\mathcal{D}$, the objective reads
\begin{equation}
\label{eq:hyper-opt}
    \theta^* = \underset {\theta \in \boldsymbol\Theta}{\arg\min} \mathbb{E}_{(\mathbf{X},\mathbf{X}_*)\sim \mathcal{D}} \mathbf{V}(\Lambda, \phi_{\theta}, \mathbf{X}, \mathbf{X}_*),
\end{equation}
where $\mathbf{V}(\Lambda, \phi_{\theta}, \mathbf{X}, \mathbf{X}_*)$ is the validation protocol which \review{measures the loss $\Lambda$} of the model $\phi_{\theta}$ with hyperparameters $\theta$ when trained on the dataset $\mathbf{X}$ and tested on dataset $\mathbf{X}_*$. In practice, we have a finite set of samples thus we compute the approximate expectation. Popular choices of the validation protocol $\mathbf{V}(\cdot)$ include the holdout and cross-validation (CV) error \cite{feurer2019hyperparameter, bischl2012resampling}. 

In this work, we found that certain DR methods are competitive with default parameters and do not benefit much from parameter tuning. These are typically simpler methods like kernel-PCA or DMAP which tend to have a small number of hyperparameters. However, for more complex models (e.g., ANNs) parameter tuning has the potential to boost the model performance compared to using default parameters. In this work, hyperparameter values are chosen based on the predictive ability of the surrogate model and not solely on the performance of the DR method (e.g., to what extent the local structure is preserved or how small the error is between original samples and inverse-mapped samples). We use default parameters for methods which do not benefit much from parameter tuning and we manually select after experimentation hyperparameter values for more complex DR methods such as AE or WAE. Therefore, we do not expect that implementing expensive optimization would significantly alter the comparison results or the drawn conclusions. Nevertheless, optimization can be straightforwardly integrated into the proposed framework for a more systematic experimentation and model selection.

\section{Numerical examples}
\label{S:Examples}

We compare the performance of the thirteen  manifold learning methods for high-dimensional UQ on the following three applications: (i) the 1D stochastic Poisson's equation, (ii) a stochastic elliptic PDE modeling a 2D diffusion process, and (iii) a 2D stochastic time-dependent PDE describing an autocatalytic reaction. For each application, we compare the performance of each method and provide numerical evidence in terms of generalization error and computational cost which consists of the CPU time for both the DR and surrogate modeling tasks. All analyses in this paper were carried out on a 2.8 GHz Quad-Core Intel Core i7 X CPU.

\subsection{1D stochastic Poisson's equation}

Consider the following one-dimensional stochastic Poisson's equation with homogeneous boundary conditions
\begin{equation}
\begin{split}
    -\odv[2]{}{x}u &= f(x;\omega), \ \ x \in [-1,1] \ \text{and} \ \omega \in \Omega, \\
    u(-1) &=u(1)=0
\end{split}
\label{eq:A1-poisson}
\end{equation}
where $\Omega$ is the random space and the forcing term is given by the following Gaussian random process
\begin{equation}
    f(x;\omega) \sim \mathcal{GP}(f_0(x), \text{Cov}(x,x'))
\label{eq:A1-forcing-term}
\end{equation}
with mean function and covariance function

\begin{equation}
\begin{split}
    f_0(x) &= 0.1 \sin (\pi x) \\
    \text{Cov}(x,x') &= \sigma^2 \exp \bigg( - \frac{(x-x')^2}{l_c^2} \bigg),
\end{split}
\label{eq:A1-mean-cov}
\end{equation}
where $\sigma = \sqrt{20}$ is the standard deviation and $l_c =0.2$ the correlation length.

\begin{figure}[ht!]
\begin{center}
\includegraphics[width=0.9\textwidth]{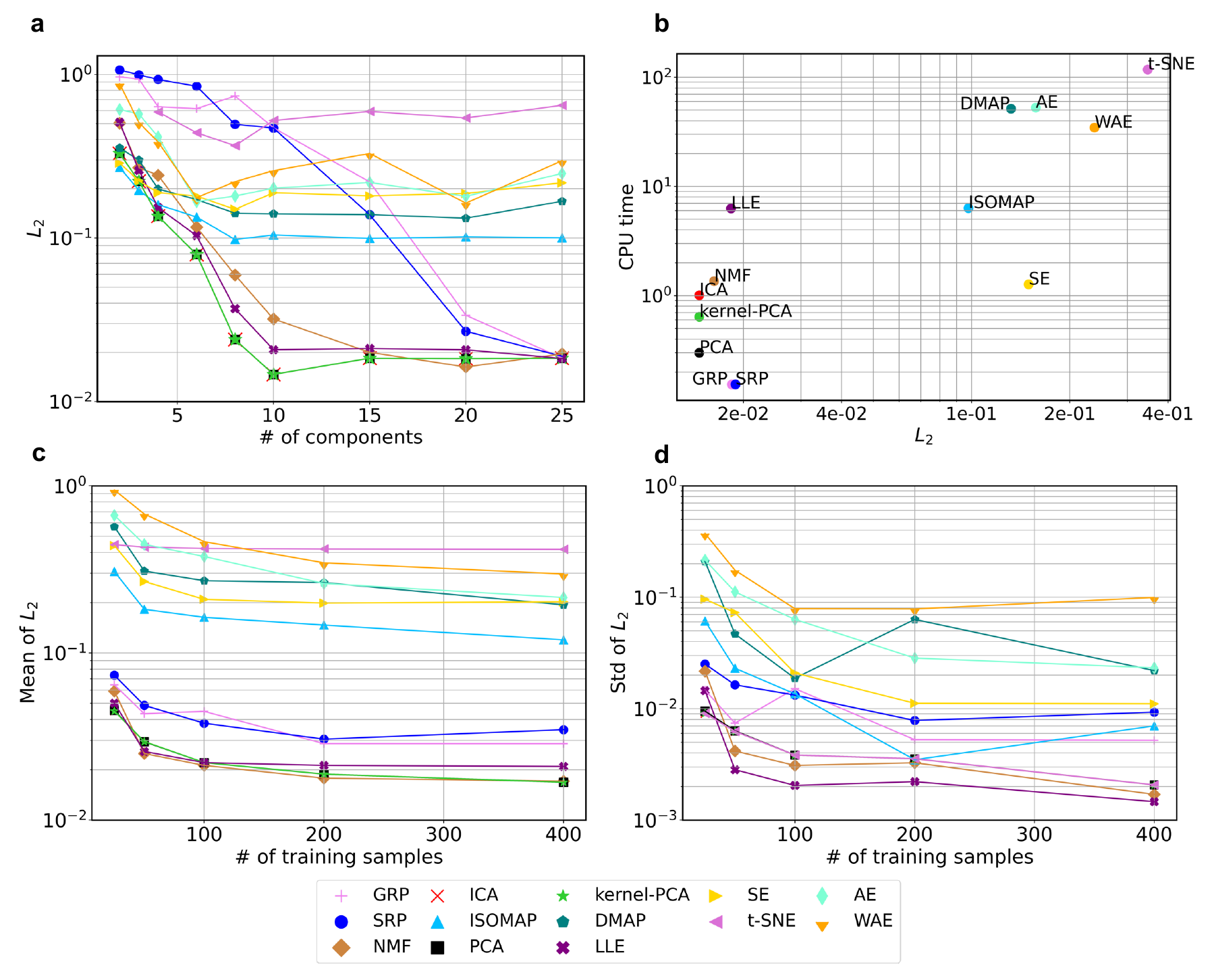}
\caption{\review{Comparison of the accuracy of the thirteen DR methods as a function of (a) the number of reduced components and (b) the CPU time of the trained PCE surrogate evaluated on the testing set of $N^*=1000$ data. Furthermore, the mean (c) and standard deviation (d) of the relative $L_2$ error are shown for $d=20$ as a function of the number of training samples based on repeated runs with different datasets. }}
\label{fig:AP1-comparison}
\end{center}
\end{figure}

The proposed approach is employed to learn the mapping from $f(x;\omega)$ to the solution $u(x;\omega)$. We discretize the 1D domain $[-1,1]$ into $t=1024$ points and generate $N=1000$ realizations of the Gaussian stochastic process, $\{ f_i = f(x;\omega) \}_{i=1}^{N}$ using the Karhunen Lo\'eve expansion (KLE) as
\begin{equation}
    f(x) = f_0(x)+\sum_{i=1}^{N_t} \sqrt{\lambda_i}\theta_i f_i(x),
\label{eq:kle_gaussian_formula}
\end{equation}
where $\lambda_i$ and $f_i(x)$ are the eigenvalues and eigenvectors respectively of the covariance function $\text{Cov}(x,x')$ determined by solving the homogeneous Fredholm integral eigenvalue problem and $\theta_i$ are iid $N(0,1)$ random variables \cite{huang2001convergence}. In the KLE, we retain $N_t = 1024$ (the full rank of the expansion) which is intentionally set to a conservatively large number to allow us to identify the intrinsic dimension through DR. \review{Next, we translate the Gaussian stochastic process generated in Eq.~\eqref{eq:kle_gaussian_formula} to a non-Gaussian stochastic process \cite{Grigoriu1998} as follows }
\begin{equation}
    f^{\text{NG}}(x) = f_0(x) + F_{\text{NG}}^{-1}\Bigg(\Phi\Bigg(\frac{\sum_{i=1}^{N_t} \sqrt{\lambda_i}\theta_i f_i(x)}{\sigma}\Bigg)\Bigg)
\end{equation}
\review{where $F_{\text{NG}}^{-1}(\cdot)$ represents the inverse cumulative distribution function of the non-Gaussian distribution and $\Phi$ represents the cdf of the standard normal distribution. In this example, we chose the non-Gaussian distribution to be the uniform distribution $U[a, b]$, where $a = -15$ and $b = 15$. The transformed non-Gaussian stochastic process is therefore given by}
\begin{equation}
\begin{aligned}
    & f^{\text{NG}}(x) = f_0(x) + a + (b-a) * \Phi\Bigg(\frac{\sum_{i=1}^{N_t} \sqrt{\lambda_i}\theta_i f_i(x)}{\sigma}\Bigg) \\
    & f^{\text{NG}}(x) = f_0(x) + a + (b-a) * \Bigg(1 + \frac{1}{2} \mathrm{erf} \Bigg(\frac{\sum_{i=1}^{N_t} \sqrt{\lambda_i}\theta_i f_i(x)}{\sqrt{2}\sigma}\Bigg)\Bigg) \\
\end{aligned}
\end{equation}
\review{where $\mathrm{erf}(z) = \frac{2}{\sqrt{\pi}} \int_0^z \exp\left(-t^2\right)\mathrm{d}t$ is the error function.}




We solve the equation with the finite difference (FD) method and generate the corresponding model responses $\{ u_i = u(x;\omega) \}_{i=1}^{N}$. Therefore we construct the training dataset of inputs $\mathbf{X} = \{\mathbf{x}_1,..,\mathbf{x}_{N}\}$ where $\mathbf{x}_i \in \mathbb{R}^{1024}$ and outputs $\mathbf{Y} = \{\mathbf{y}_1,..,\mathbf{y}_{N}\}$ where $\mathbf{y}_i \in \mathbb{R}^{1024}$. To evaluate the performance of the trained model, we generate another $N^*=1000$ samples of $f(x;\omega)$ and corresponding trajectories $u(x;\omega)$ independently of the training data. We denote this dataset as $\mathbf{X}^* = \{\mathbf{x}_1,..,\mathbf{x}_{N^*}\}$, $\mathbf{Y}^* = \{\mathbf{y}_1,..,\mathbf{y}_{N^*}\}$ of inputs and model outputs respectively. The quality of the surrogate predictions are evaluated based on the metrics presented in Section \ref{error-metrics}.

We compare the performance of the trained PCE surrogate for all DR methods in Figure \ref{fig:AP1-comparison}. \review{Figures \ref{fig:AP1-comparison}(a) and \ref{fig:AP1-comparison}(b) show the relative $L_2$ error and CPU training time of the trained m-PCE where each color/symbol corresponds to one of the thirteen DR methods. We observe that the simple DR methods converge fast and with high accuracy. The optimal response in terms of relative error is observed by the PCA, k-PCA, ICA methods which resulted in similar relative errors. We note, that in this example a linear kernel has been chosen for k-PCA therefore the method essentially results in standard PCA. However, the CPU time between these methods varies significantly as shown in Figure \ref{fig:AP1-comparison}(c). Taking into account both the prediction accuracy and CPU time, PCA PCE performed optimally resulting in a relative error of $L_2=2.80 \cdot 10^{-5}$ where the PCE surrogate is constructed with $d=20$ approximately independent input random variables. Furthermore, in \ref{fig:AP1-comparison}(c),(d) we plot the mean and standard deviation of the relative $L_2$ error versus the number of training data. We observe that the best performing methods converge rapidly (with only $\sim100$ training points) to an accurate solution with small uncertainty, while more expensive methods such as AE and WAE result in a much higher mean value and error.} 

\begin{figure}[ht!]
\begin{center}
\includegraphics[width=0.9\textwidth]{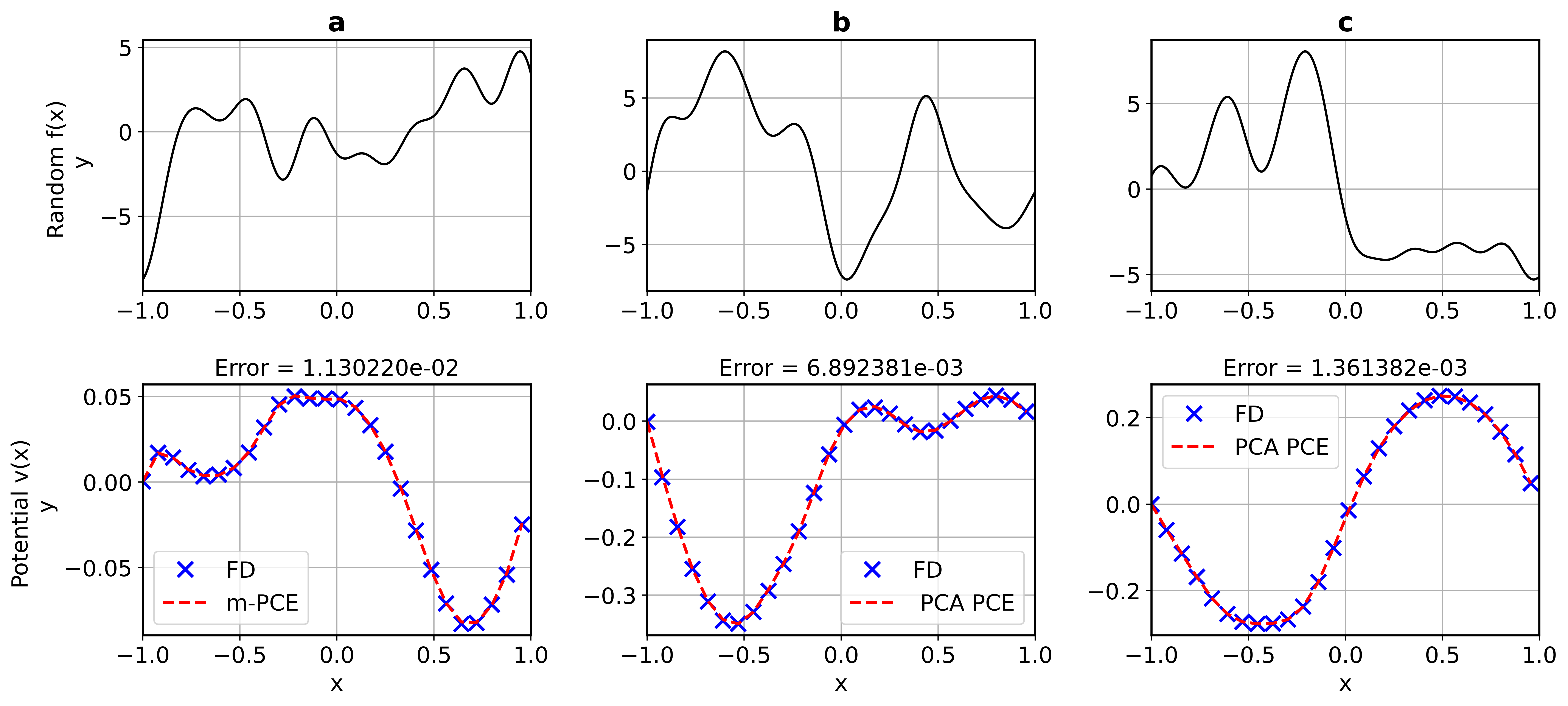}
\caption{PCA PCE prediction for three random realizations of the inputs stochastic process. Blue `x's represent the solution with the numerical solver (finite differences or FD) and red dashed curves the m-PCE response.}
\label{fig:AP1-comparison-random}
\end{center}
\end{figure}

\begin{figure}[ht!]
\begin{center}
\includegraphics[width=0.5\textwidth]{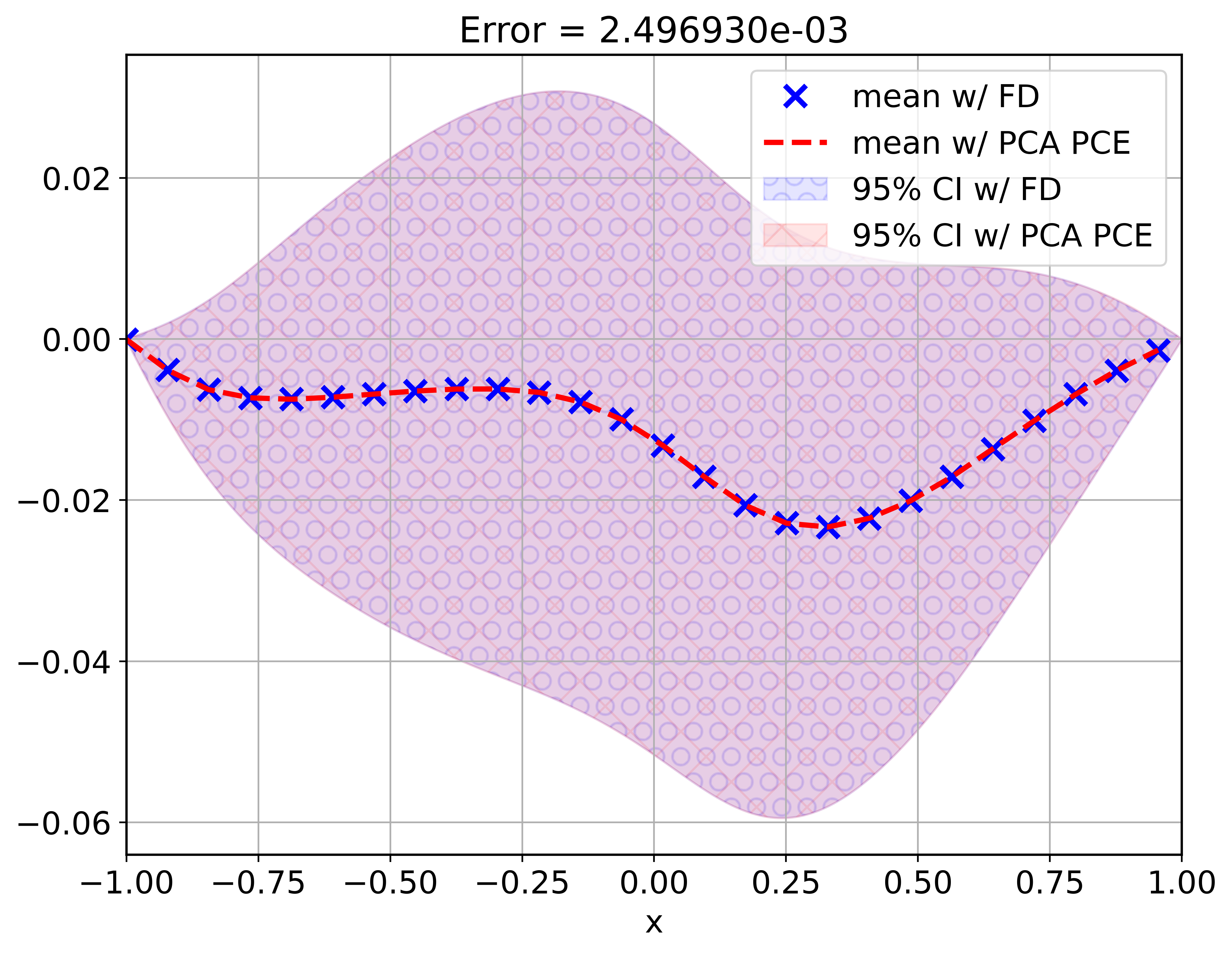}
\caption{Estimation of the mean and $95 \%$ CI of solution with FD and m-PCE.}
\label{fig:AP1-moment-estimation}
\end{center}
\end{figure}

In Figure \ref{fig:AP1-comparison-random}, the predictive accuracy of the optimal PCA PCE surrogate is illustrated. Three random realizations of the input stochastic process are generated and the surrogate response is compared with the ground truth solution computed with the FD solver. The comparison, results in scalar error measures in the order of $10^{-5}$, revealing the powerful predictive ability of the surrogate. Finally, in Figure \ref{fig:AP1-moment-estimation}, the mean response and the associated $95 \%$ CI are computed for all $N^*$ test data. We observe that the m-PCE has the ability to perform moment estimation with very high accuracy. 

\subsection{2D stochastic steady-state heat equation}
\label{AP2}
Consider the following elliptic PDE on the 2-d unit square domain which describes steady-state diffusion processes
\begin{subequations}
\begin{align}
    \nabla(D(\mathsf{x})\nabla u(\mathsf{x}))&=0, \ \forall \mathsf{x} \in [0,1]^2 \\ 
    \text{BC:} \hspace{5pt} u&=0, \hspace{2pt} \forall x = 1, \\
    u&=1, \hspace{2pt} \forall x=0,\\
    \frac{\partial u}{\partial n} &=0, \hspace{2pt} \forall y=0 \hspace{2pt} \text{and} \hspace{2pt} y=1,
\end{align}
\label{eq:A2-model}
\end{subequations}
where $\mathsf{x}=(x,y)$ represents the spatial coordinates in 2-d Euclidean space. In this example, randomness comes from the spatially varying diffusion coefficient $D(\mathsf{x})$, which is modeled and sampled from a log-normal random field such that
\begin{equation}
\label{eq:A2:log-normal}
    \log D(\mathsf{x}) \sim \mathcal{GP}(D(\mathsf{x})|\mu(\mathsf{x}), \text{Cov}(\mathsf{x},\mathsf{x}')),
\end{equation}
where $\mu(\mathsf{x})$ and $\text{Cov}(\mathsf{x},\mathsf{x}')$ are the mean and covariance functions respectively. For simplicity, we set $\mu(\mathsf{x})=0$, while the covariance matrix is given by the squared exponential kernel as
\begin{equation}
\label{eq:A2:cov}
    \text{Cov}(\mathsf{x},\mathsf{x}') = \text{exp} \Bigg( - \frac{|x - x'|}{\ell_x} - \frac{|y - y'|}{\ell_y}\Bigg),
\end{equation}
where $\ell_x$, $\ell_y$ are the correlation lengthscales along the $x$ and $y$ spatial directions, respectively.
To generate realizations of the input stochastic diffusion field we again employ the KLE with a full rank of 784 random variables, again to allow the DR to identify the intrinsic dimension. Traditionally, one would construct a surrogate model to map the input stochastic fields $D(\mathsf{x})$ to the model output $u(\mathsf{x})$ for constant values of the lengthscale parameters. Similarly to \cite{tripathy2018deep}, we aim to remove the restrictions of the lengthscales and therefore we construct a surrogate which is trained to predict the numerical solution $u(\mathsf{x})$ for novel realizations of the discretized random input fields for arbitrary lengthscale values. 

\begin{figure}[ht!]
\begin{center}
\includegraphics[width=0.6\textwidth]{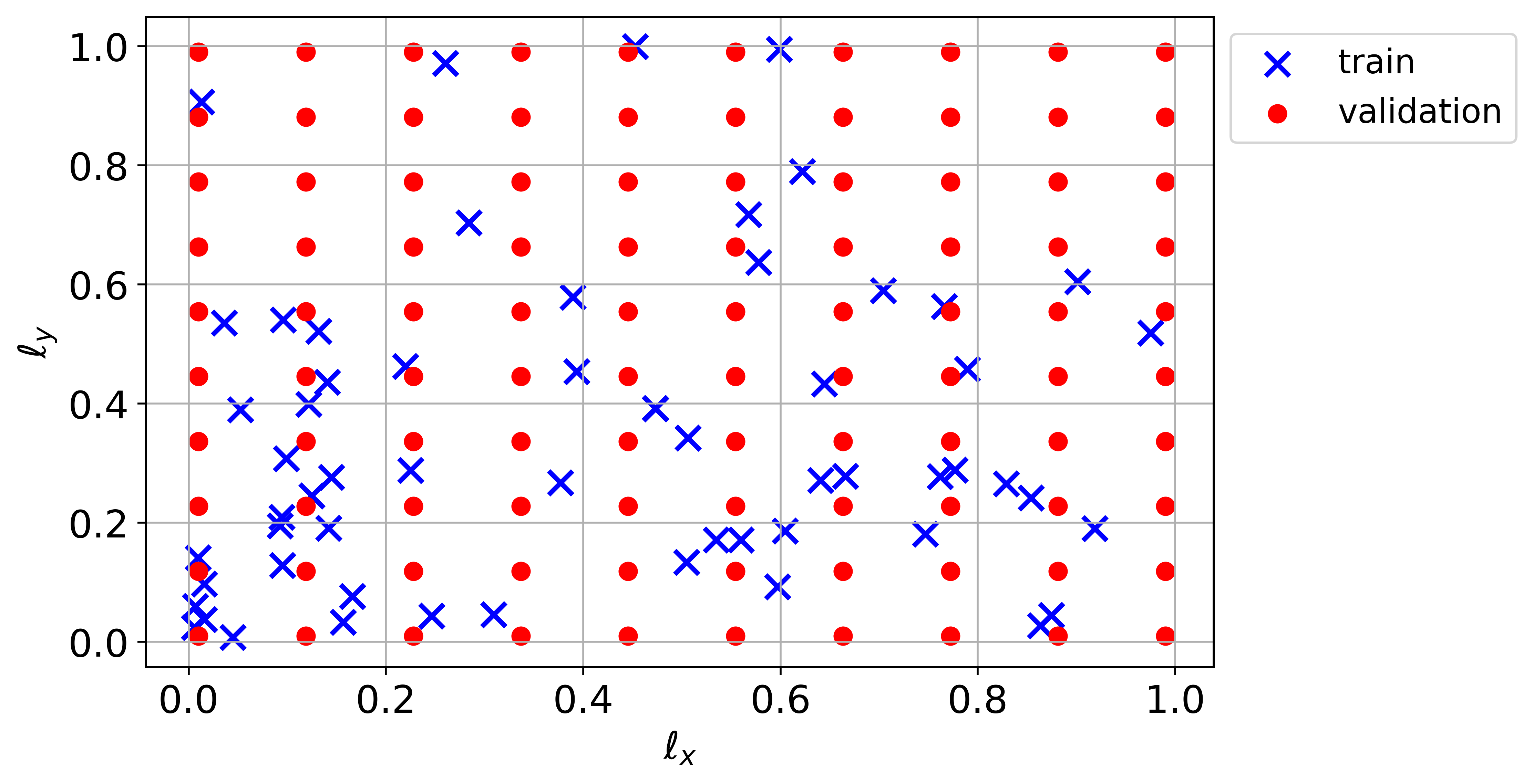}
\caption{Randomly generated lengthscale pairs with LHS for the training dataset (blue) and uniform lengthscale pairs (red) for the validation dataset.}
\label{fig:AP2-lengthscales}
\end{center}
\end{figure}

The simulation takes place in a square domain $\Omega= [0,1] \times [0,1]$, discretized into $32\times32=1024$ volume cells and the PDE is solved with the finite volume method (FVM). To construct the training dataset, we first obtain the design of lengthscale pairs $\mathbf{L} = (\ell_x,\ell_y)$. We generate $n=60$ lengthscale pairs using the sampling algorithm proposed in \cite{tripathy2018deep} to bias samples toward lower length-scales and for each pair, $N=70$ realizations of the input stochastic field. For these $n \cdot N = 4200$ total realizations we solve the forward model and split the dataset into $N=2000$ training and $N^*=2200$ test data. We denote the input train data with $\mathbf{X} = \{\mathbf{x}_1,..,\mathbf{x}_{N}\}$ where $\mathbf{x}_i \in \mathbb{R}^{1024}$ and the output solutions with $\mathbf{Y} = \{\mathbf{y}_1,..,\mathbf{y}_{N}\}$ where $\mathbf{y}_i \in \mathbb{R}^{1024}$. In Figure \ref{fig:AP2-lengthscales}, the lengthscale pairs are shown. We also generate input fields and model solutions for a set of $\hat{n} = 100$ lengthscale pairs uniformly distributed in the square spatial domain as an out-of-distribution (OOD) validation set to test generalizability, also shown in Figure \ref{fig:AP2-lengthscales}, with $\hat{N}=50$ samples for each pair. The generalization ability of the trained surrogate will be tested for this OOD validation dataset $(\hat{\mathbf{X}},\hat{\mathbf{Y})}$ which is not used during the training of the model.

\subsubsection{Prediction for arbitrary lengthscales}

\begin{figure}[ht!]
\begin{center}
\includegraphics[width=0.9\textwidth]{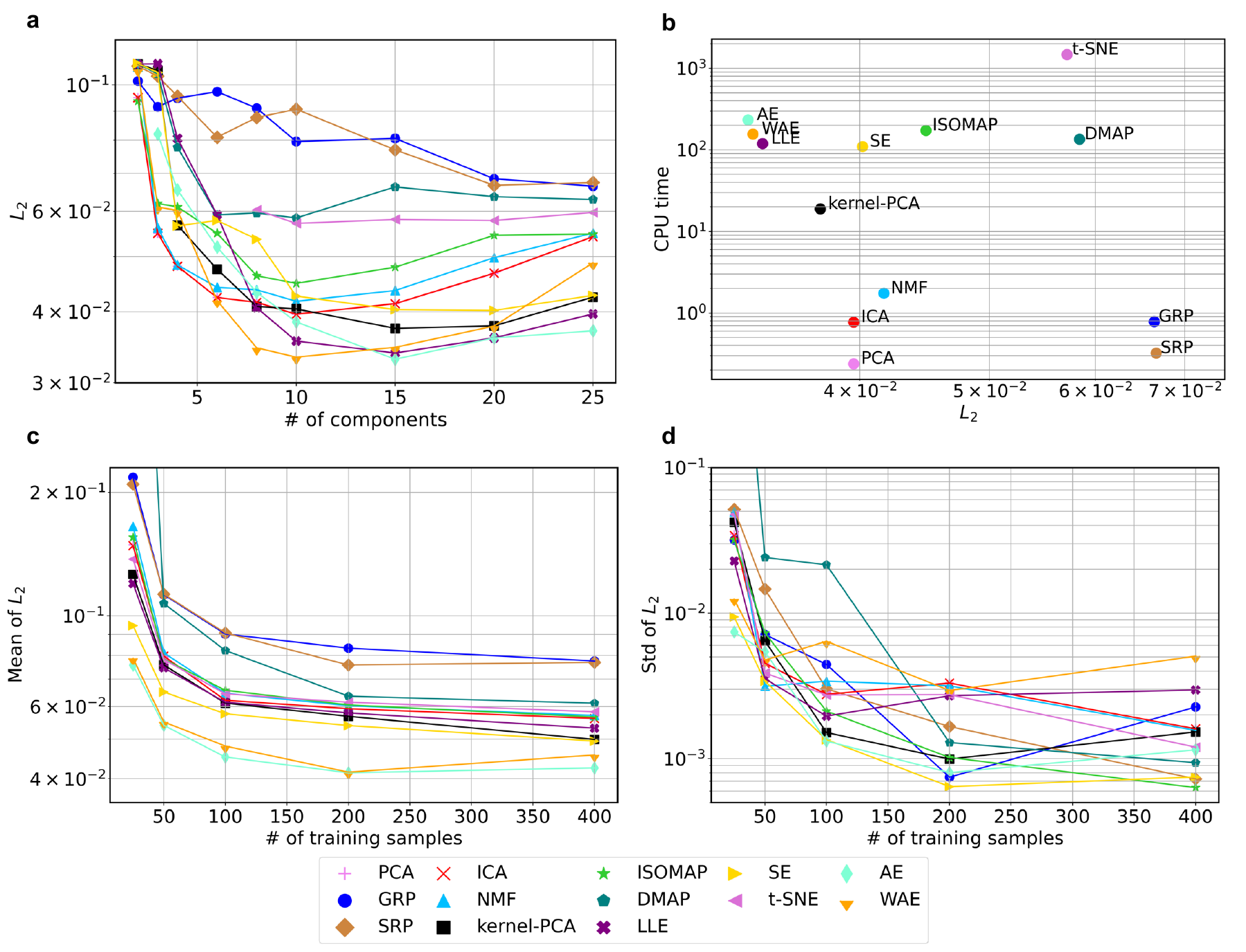}
\caption{\review{Comparison of the accuracy of the thirteen DR methods as a function of the (a) number of reduced components and (b) the CPU time of the trained PCE surrogate evaluated on the dataset $(\hat{\mathbf{X}},\hat{\mathbf{Y})}$. Furthermore, the mean (c) and standard deviation (d) of the relative $L_2$ error are shown for $d=15$ as a function of the number of training samples based on repeated runs with different datasets. }}
\label{fig:AP2-comparison}
\end{center}
\end{figure}

\begin{figure}[ht!]
\begin{center}
\includegraphics[width=0.85\textwidth]{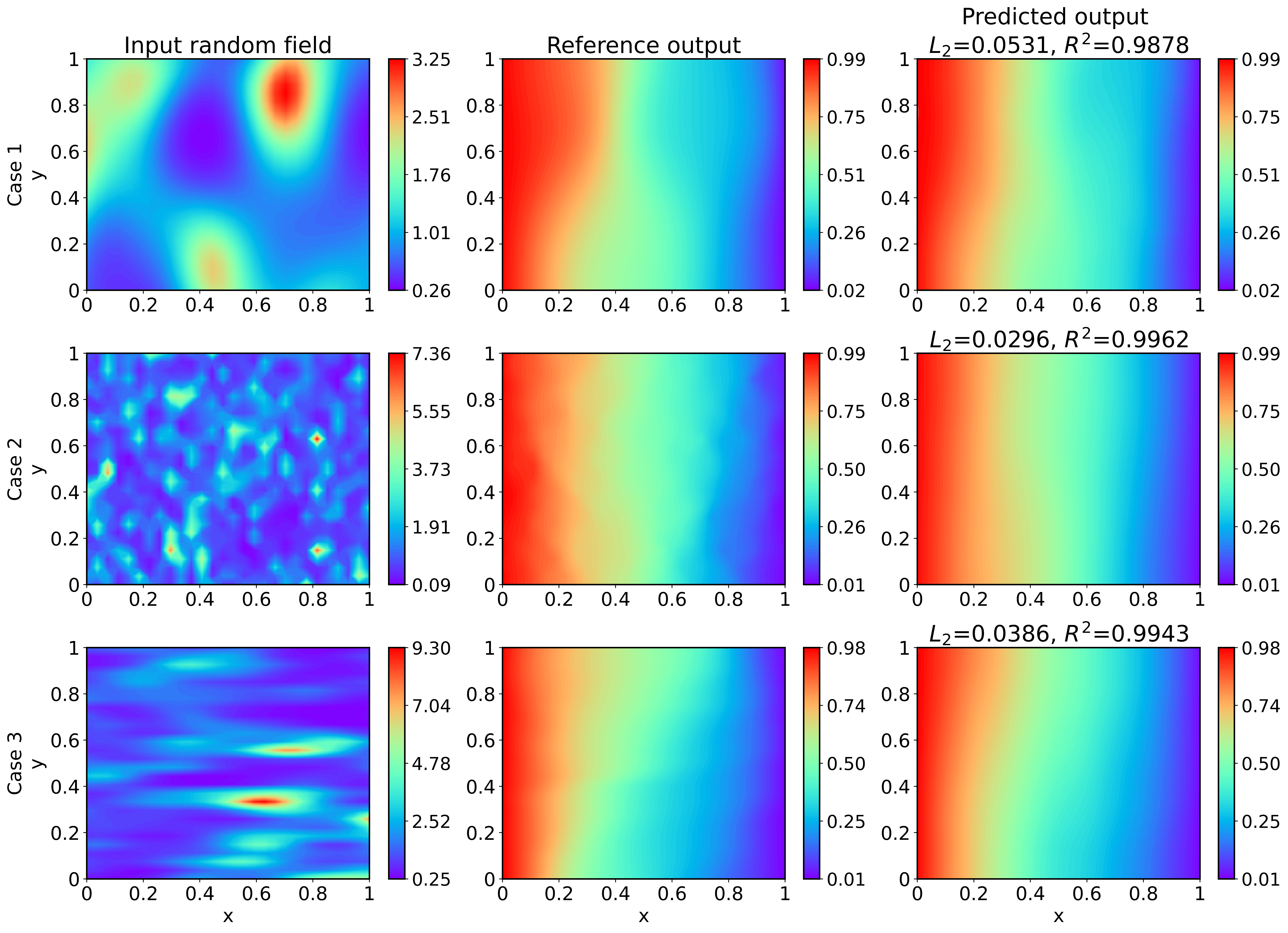}
\caption{AE PCE prediction for three random realizations of the input diffusion field $D(\mathsf{x})$.}
\label{fig:AP2-ae-pce-comparison}
\end{center}
\end{figure}

\begin{figure}[ht!]
\begin{center}
\includegraphics[width=0.8\textwidth]{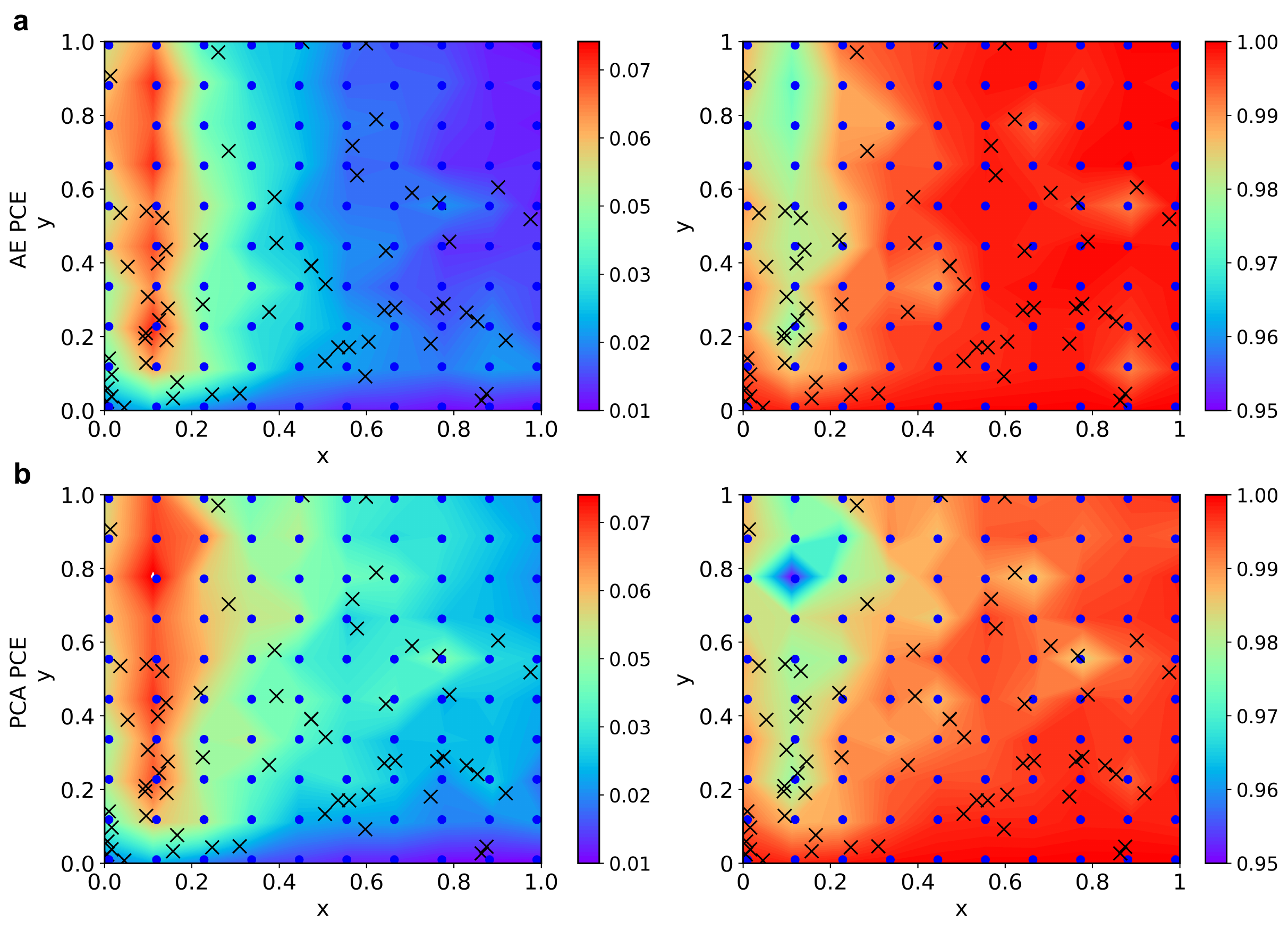}
\caption{Average of the (left column) relative $L_2$ errors and (right column) $R^2$ scores for each uniform lengthscale pair. `x' markers represent the training lengthscale pairs generated with LHS and solid `o' markers the uniform lengthscale pairs used for validating the (a) AE PCE and (b) PCA PCE surrogate.}
\label{fig:AP2-ae-pce-error}
\end{center}
\end{figure}

A simple PCA on the $1024$-dimensional input data reveals that $15$ dimensions account for $\approx 87 \% $ of the variance of the data. We compare the performance of the trained PCE surrogate in Figure \ref{fig:AP2-comparison} evaluated on the validation dataset of uniformly distributed lengthscales $(\hat{\mathbf{X}},\hat{\mathbf{Y})}$. \review{Figures \ref{fig:AP2-comparison}(a) shows the relative $L_2$ error of the trained m-PCE as a function of the reduced space dimension (number of components) where each color corresponds to one of the thirteen DR methods.} We observe that the WAE, AE and LLE performed best in terms of predictive accuracy, while GRP and SRP resulted in the largest error. The optimal response in terms of relative error is observed by the AE PCE and in terms of CPU time by the PCA PCE, where the surrogate is constructed with a diffusion input field modeled as a nonlinear combination of $d=15$ and $d=10$ independent standard normal random variables respectively and a maximum polynomial degree of $s_{max}=2$ for both cases. The resulting average relative error is measured as $L_2 = 3.30\cdot 10^{-2}$ and $L_2 = 3.95\cdot 10^{-2}$ respectively. Figure \ref{fig:AP2-comparison}(b), shows the CPU time versus the relative $L_2$ error. Although AE, WAE and LLE performed best, their computational cost is significantly higher (two orders of magnitude). ICA, PCA and NMF methods provide a much faster alternative while still exhibiting a small validation error $L_2 \approx 4\cdot 10^{-2}$. \review{Finally, in Figure \ref{fig:AP2-comparison}(c),(d) the surrogate is trained for different training datasets to obtain confidence intervals for the predictions. The mean and standard deviation are shown as a function of the number of training samples. We observe that the accuracy for all methods converges quickly (within $\approx 200$ training samples) with small uncertainty, thus small datasets suffice.}

In Figure \ref{fig:AP2-ae-pce-comparison}, the results of the optimal surrogate in terms of predictive accuracy (i.e., AE PCE) are shown for three random test realizations of the input diffusion field $D(\mathsf{x})$. These three fields, demonstrate the lengthscale variability which renders the input dataset challenging for the construction of the surrogate model. As expected, the error increases for finer lengthscales in both spatial dimensions which results in input fields with more variation and thus more intricate underlying structure. The average of the relative $L_2$ errors and the $R^2$ scores for each uniform lengthscale pair is calculated for both the AE PCE (highest accuracy) and PCA PCE (highest speed) and shown in Figure \ref{fig:AP2-ae-pce-error}. We observe that the surrogate performs very well even for input fields with lengthscale pairs that have not been included in the training of the model. Based on these results, we suggest that one should sample more pairs that corresponds to small $\ell_x$ values since in this zone we compute the highest relative errors. Comparing these plots with the equivalent ones from a similar approach based on the training of DNNs for both the DR and surrogate modeling tasks, known as the `Deep UQ' proposed in \cite{tripathy2018deep}, we observe that the proposed AE PCE approach results in comparable response in terms of predictive accuracy and generalization, with less training data and CPU time. Therefore, the m-PCE can provide a cost-effective approach for constructing surrogates with less trainable parameters, when small data is available with limited computational resources.

\subsubsection{Uncertainty propagation}

\begin{figure}[ht!]
\begin{center}
\includegraphics[width=0.8\textwidth]{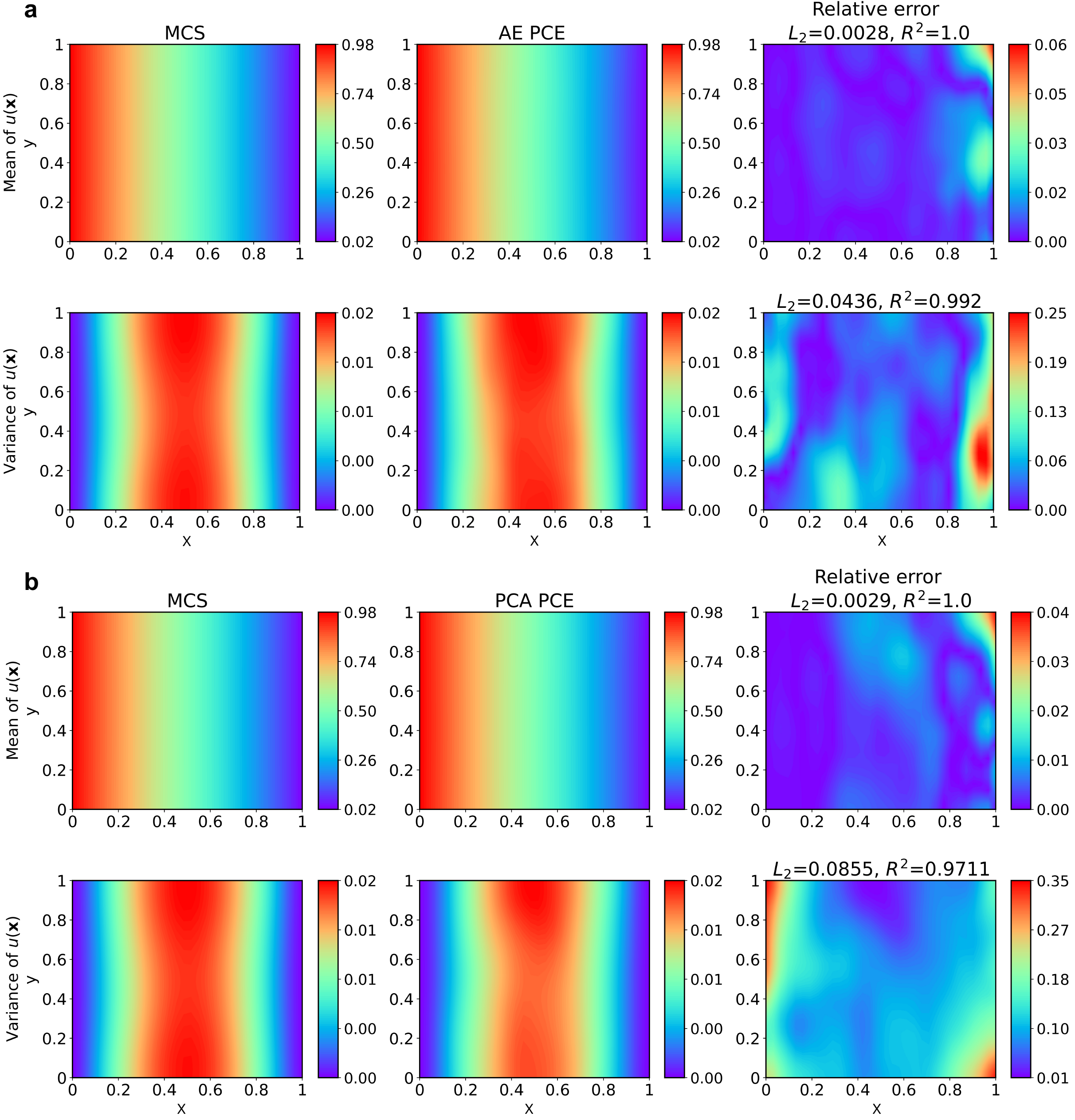}
\caption{Mean and variance field of PDE solution $u(\mathsf{x})$ for $\ell_x = 0.15$ and $\ell_y=0.25$. The fields are computed with MCS (first column) and (a) AE PCE and (b) PCA PCE (second column) for $10^4$ samples. Third column shows the relative error as a field and also the scalar relative $L_2$ error (average) and $R^2$ score.}
\label{fig:AP2-ae-moment-estimation}
\end{center}
\end{figure}

The trained m-PCE surrogate has been generalized for arbitrary lengthscales and will be now used for uncertainty propagation with fixed lengthscale values $\ell_x = 0.15$ and $\ell_y=0.25$. We generate a large number of input samples i.e., $10^4$ and compute with both MC simulation and the proposed surrogate the mean and variance fields of the solution $u(\mathsf{x})$. In addition we compute the probability density of the QoI at two fixed points in space: $\mathsf{x}_1 = (0.323,0.645)$ and $\mathsf{x}_2 = (0.806,0.258)$. 

\begin{figure}[ht!]
\begin{center}
\includegraphics[width=0.95\textwidth]{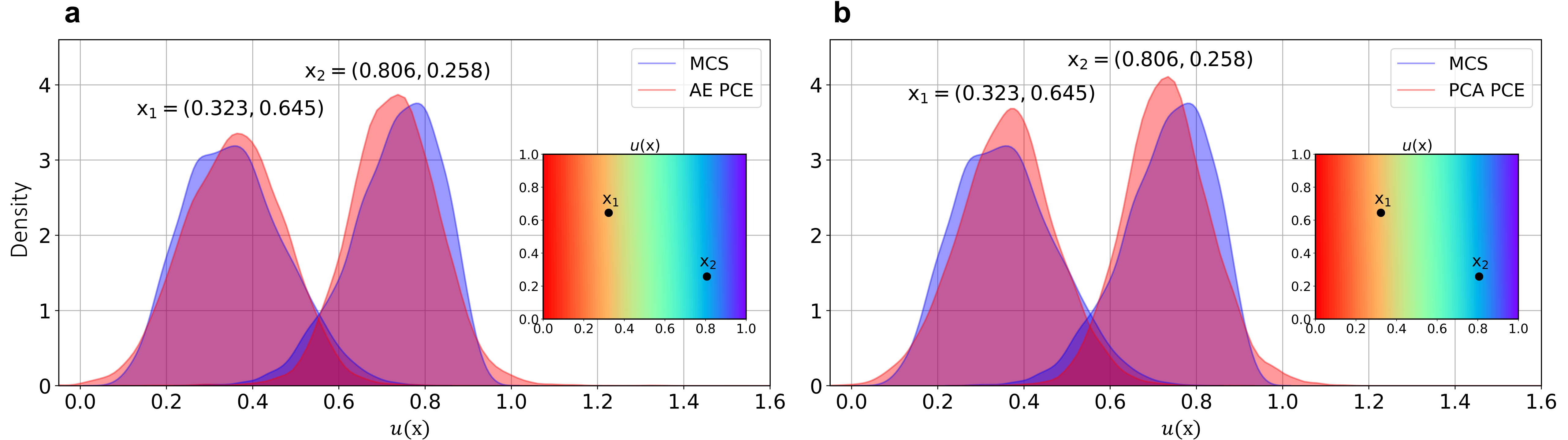}
\caption{Density of PDE solution $u(\mathsf{x})$ for $\mathsf{x}_1 = (0.323,0.645)$ and $\mathsf{x}_2 = (0.806,0.258)$ with MCS (blue) and (a) AE PCE and (b) PCA PCE (red).}
\label{fig:AP2-ae-pdf}
\end{center}
\end{figure}

We compare the mean and variance fields computed with MCS and m-PCE in Figure \ref{fig:AP2-ae-moment-estimation}. The relative error between the fields as well as the average relative $L_2$ error and $R^2$ score are computed. All error measures show a very good match between the reference and predicted mean fields. The variance fields have higher relative error, partially attributed to their small magnitude, which appears in the denominator of the $L_2$ error metric. However, the scalar error measures remain fairly low. We note that smaller lengthscales $\ell_x, \ell_y$ will likely result in higher errors, however generating more training samples for lengthscale pairs in these regions will enhance the performance and accuracy of the surrogate. Finally, Figure \ref{fig:AP2-ae-pdf} shows the comparison between the probability density of the solution $u(\mathsf{x})$ for two spatial points computed with MCS and both AE PCE and PCA PCE. We observe that even if some accuracy is sacrificed over cost, m-PCE is able to capture the density of the solution to a satisfactory degree. 

\subsection{2D time-dependent stochastic Brusselator model}
\label{AP3}

Finally, consider the stochastic Brusselator model, describing the dynamics for a type of an autocatalytic oscillating chemical reaction \cite{vaidyanathan2015dynamics}. 
The Brusselator model is characterized by the following reactions
\begin{subequations}
\begin{align}
    A & \xrightarrow{k_1} X \quad \\ 
    B + X  & \xrightarrow{k_2} Y + D \\
    2X + Y & \xrightarrow{k_3} 3X \\
    X & \xrightarrow{k_4} E
\end{align}
\label{eq:A3-reaction}
\end{subequations}
where a reactant A is converted to a final product $E$, through four steps and four additional species, $X,B,Y$ and $D$. In Eq.~\eqref{eq:A3-reaction}(b and c), bimolecular and autocatalytic trimolecular reactions are shown respectively. Based on this reaction, species $E$ is resulted from species $X$ which in turn results from Eq.~\eqref{eq:A3-reaction}(a and c). We consider that A and B are in vast excess and thus can be modeled at constant concentration. Through the law of mass action the Brusselator system can be described by the following model
\begin{equation}
\begin{split}
    \frac{\partial u}{\partial t} &= D_0 \nabla^2 u + a - (1+b)u + vu^2 \\
    \frac{\partial v}{\partial t} &= D_1 \nabla^2 v + bu - vu^2, \hspace{10pt} \mathsf{x} \in [0,1]^2, \ t \in [0,1],
\end{split} 
\label{eq:A3-model}
\end{equation}
where $\mathsf{x}=(x,y)$ are the spatial coordinates, $D_0, D_1$ represent the diffusion coefficients, $a=\{A\}, b=\{B\}$ are constant concentrations and $u=\{X\}, v=\{Y\}$ represent the concentrations of reactant species $X,Y$. Depending on the values of the parameters the Brusselator model can exhibit a limit cycle, a Hopf bifurcation and also a chaotic behaviour \cite{vaidyanathan2015dynamics}. We choose $a=1$, $b=3$, $D_0=1$ and $D_1=0.1$ for which the system is in an unstable regime approaching a limit cycle which forces the system to oscillate after sufficient time. 

\begin{figure}[ht!]
\begin{center}
\includegraphics[width=0.93\textwidth]{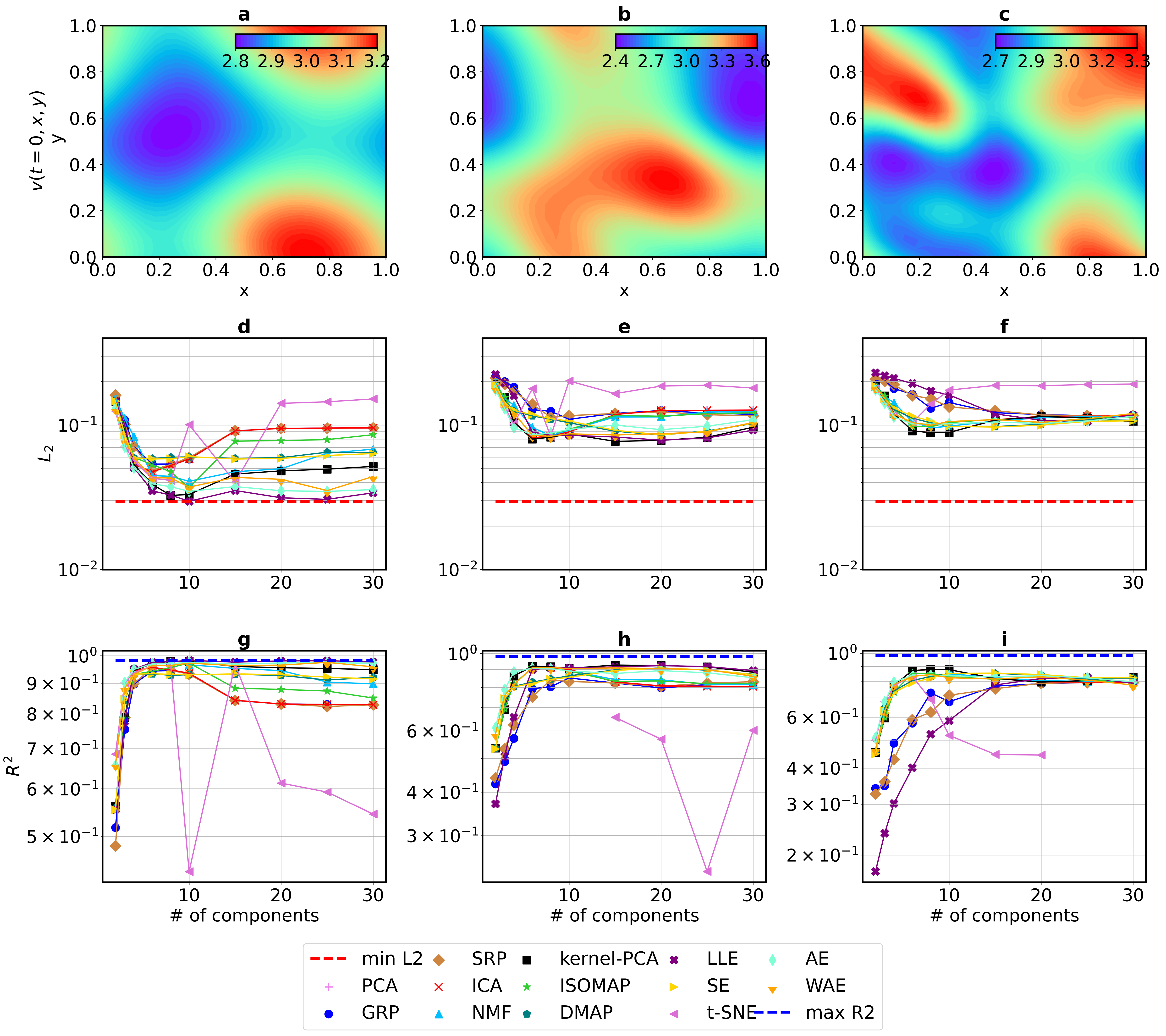}
\caption{Comparison of surrogate response for thirteen DR methods for input random fields of increasing complexity. Subplots a,b,c, represent realizations of Gaussian random fields of $v(t=0,x,y)$ with the SRM method for $(\alpha_1^{(1)}, \alpha_2^{(1)})= (3\cdot10^3,25), (\alpha_1^{(2)}, \alpha_2^{(2)})= (750,14)$ and $(\alpha_1^{(3)}, \alpha_2^{(3)})= (50,10)$ respectively and the plots below show the respective errors. Red and blue dashed lines represent the optimal relative error and $R^2$ score respectively from all three cases to serve as a reference value and are shown in all subplots (d-i).}
\label{fig:AP3-overall-comparison}
\end{center}
\end{figure}

\begin{figure}[ht!]
\begin{center}
\includegraphics[width=0.85\textwidth]{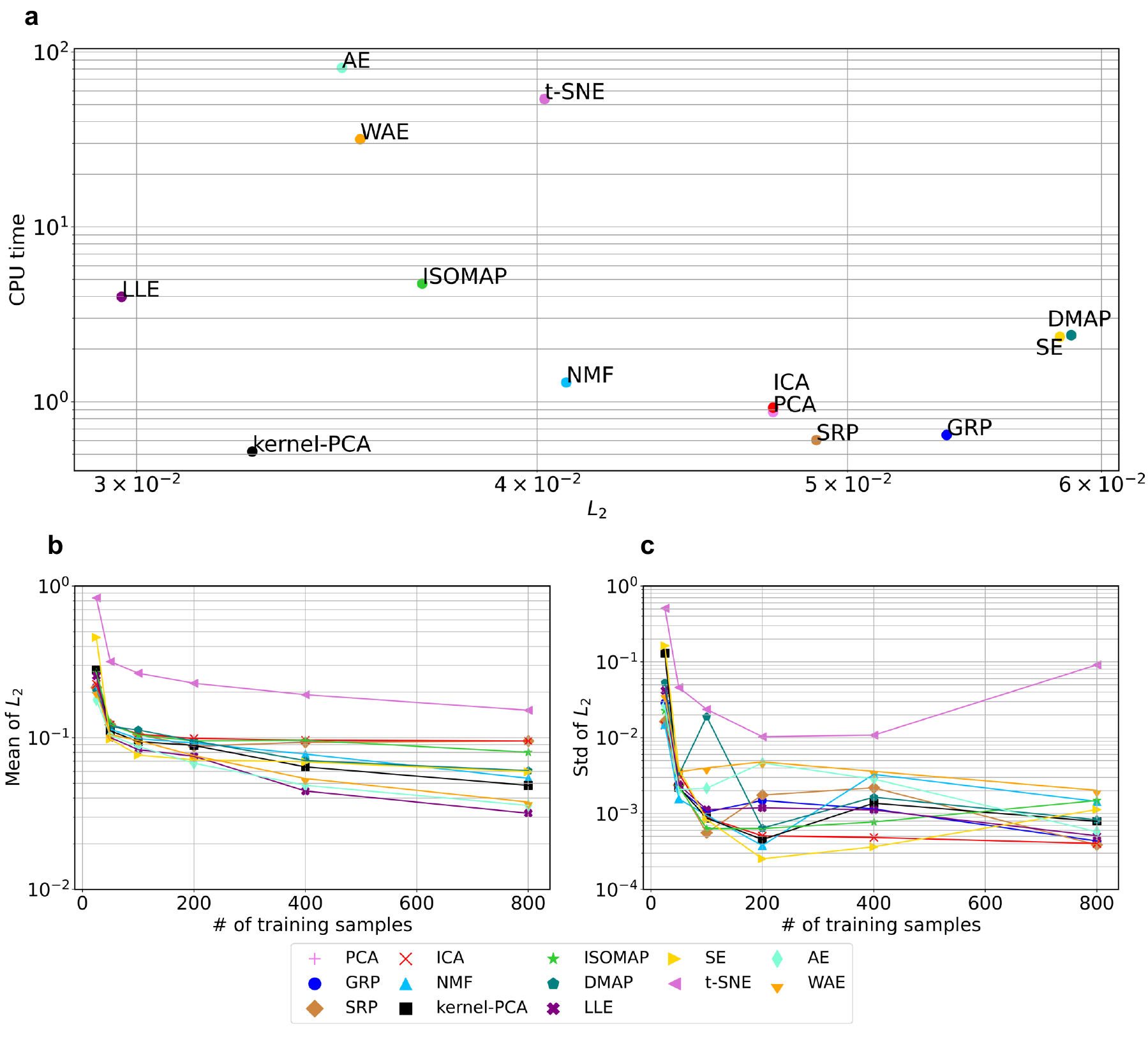}
\caption{\review{(a) CPU time versus relative error for input random realizations generated with SRM with $(\alpha_1^{(1)}, \alpha_2^{(1)})= (3\times10^3,25)$. (b),(c) present the mean and standard deviation of relative $L_2$ error respectively for $d=20$ retained components as a function of the number of training samples.}}
\label{fig:AP3-time-vs-error}
\end{center}
\end{figure}

\begin{figure}[ht!]
\begin{center}
\includegraphics[width=0.75\textwidth]{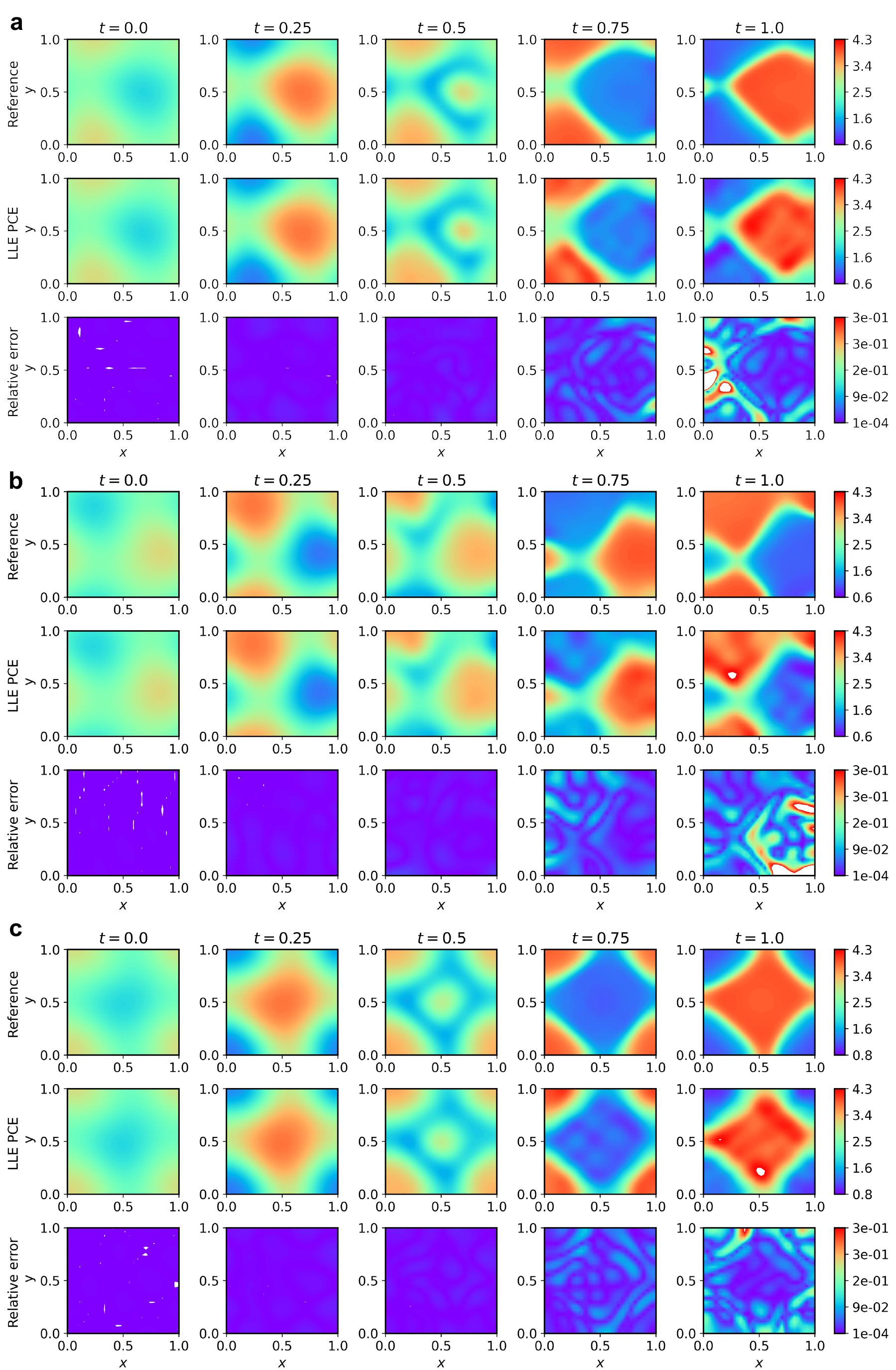}
\caption{\review{Comparison of the time evolution of $v(t,x,y)$ between reference and LLE PCE response for three input realizations generated with SRM with $(\alpha_1^{(1)}, \alpha_2^{(1)})= (3\times10^3,25)$. The reference solution and the relative error are shown for three test cases (a),(b),(c), and five time steps, i.e., $t=(0,0.25,0.5,0.75,1)$}.}
\label{fig:AP3-time-evolution}
\end{center}
\end{figure}

The simulation takes place in a square domain $\Omega= [0,1] \times [0,1]$, discretized with $28\times28=784$ mesh points. The proposed approach is employed to learn the mapping from $v(t=0,x,y)$ to the solution $v(t,x,y)$. To construct the training dataset, we generate realizations of the input Gaussian stochastic field which represents the spatially varying initial conditions of quantity $v$. We employ the spectral representation method (SRM) \cite{shinozuka1996simulation} which expands the stochastic process as a Fourier-type expansion of cosines as
\begin{equation}
\begin{aligned}
\label{eq:A3:SRM}
A(x,y) = \sqrt{2} \sum_{i=1}^{M} \sum_{j=1}^{M} \{ \sqrt{2S(\kappa_{1i}, \kappa_{2j})\Delta \kappa_1 \Delta \kappa_2} \cos{(\kappa_{1i} x + \kappa_{2j} y + \phi^{(1)}_{ij})} + \\  + \sqrt{2S(\kappa_{1i}, -\kappa_{2j})\Delta \kappa_1 \Delta \kappa_2} \cos{(\kappa_{1i} x - \kappa_{2j} y + \phi^{(2)}_{ij})} \},
\end{aligned}
\end{equation}
where $S(\kappa_1, \kappa_2)$ is the discretized power spectrum, taking the following form
\begin{equation}
    S(\kappa_1,\kappa_2)= \alpha_2 (\kappa_1^2 + \kappa_2^2) e^{-\alpha_1 \sqrt{\kappa_1^2 + \kappa_2^2}}
\end{equation}
$\Delta \kappa_1 = \Delta \kappa_2 =0.01$ is the wavenumber discretization, $\kappa_{1u}=\kappa_{2u}=1.28 \text{rad} / \text{sec}$ the upper cutoff wavenumber which yields $M=128$, and thus a total of 32,768 random phase angles $\phi_{ij}$ that are uniformly distributed in $[0, 2\pi]$. The two parameters of the discretized spectrum $(\alpha_1, \alpha_2)$ dictate the complexity and thus the intrinsic dimensionality of the generated random fields. We are considering the following three cases: $(\alpha_1^{(1)}, \alpha_2^{(1)})= (3\cdot10^3,25), (\alpha_1^{(2)}, \alpha_2^{(2)})= (750,14)$ and $(\alpha_1^{(3)}, \alpha_2^{(3)})= (50,10)$. Using SRM in UQpy \cite{olivier2020uqpy}, we generate a dataset of $N=800$ input random fields $\mathbf{X} = \{\mathbf{x}_1,..,\mathbf{x}_{N}\}$ where $\mathbf{x}_i \in \mathbb{R}^{784}$ for each one of the three cases. The initial condition of $u$ is represented by a constant field with a value $u(t=0,x,y)=a$. By solving the PDE system in the time interval $t=[0,1]$ for $\delta t=10^{-2}$, we measure the spatial behaviour of $v(\mathsf{x})$ at ten different time points, for the $N$ input stochastic field realizations. We solve the system with finite differences and vectorize solutions thus obtaining the output response matrices as $\mathbf{Y} = \{\mathbf{y}_1,..,\mathbf{y}_{N}\}$ where $\mathbf{y}_i \in \mathbb{R}^{7840}$. We collect another $N^*=1000$ input samples and corresponding model outputs to evaluate the performance of the surrogate model.

We note that constructing a surrogate model for input random fields generated with the SRM method may be more challenging than for fields generate using the KLE because the SRM provides a much less compact representation of the stochastic process and therefore the intrinsic dimensionality of the fields may be significantly higher. The computational complexity of the PCE surrogate training increases exponentially with the dimensionality of inputs, therefore the goal is to keep this number as small as possible. Furthermore, in this application we aim to predict the full time evolving solution of the Brusselator model and not a single solution snapshot, which introduces an additional challenge to the construction of an accurate surrogate. 

In Figure \ref{fig:AP3-overall-comparison}, we compare the predictive accuracy of the m-PCE surrogate for input random fields generated with the SRM method for the three cases discussed above. Each column shows a representative input random field, the relative error and $R^2$ score for all thirteen DR methods as a function of the embedding dimensionality (number of components). \review{Figure \ref{fig:AP3-time-vs-error} (a), shows the $L_2$ error vs. CPU time for each of the 13 methods for the case of $(\alpha_1^{(1)}, \alpha_2^{(1)})= (3\times10^3,25)$. These results show that LLE performs best when considering random fields with large lengthscale value (higher decay parameter $\alpha_2$) resulting in an average relative error of $L_2 = 2.97 \cdot 10^{-2}$. However, k-PCA seems to perform better when considering the CPU time, and having only slightly higher $L_2$ error. Figure \ref{fig:AP3-time-vs-error} (b),(c) show the mean and standard deviation of relative $L_2$ error for all 13 methods for $d=20$. We observe a fast convergence of the error with respect to the number of training samples with small uncertainty for most methods. However, certain methods such as LLE, AE and WAE require more data to reach convergence.} Furthermore, from Figure \ref{fig:AP3-overall-comparison}, PCA, k-PCA and ICA seem to outperform LLE in terms of predictive accuracy of the surrogate when the lengthscale of random fields is smaller and thus the complexity is higher. The accuracy of the PCE surrogate fluctuates significantly when the t-SNE method is employed, which is most likely attributed to convergence issues related to non-convex optimization. Interestingly, the surrogate accuracy for certain DR methods, reaches a minimum point and subsequently increases. One possible explanation is that the DR method has identified the intrinsic dimensionality of input data, therefore adding more components results in including redundant and noisy features. Another explanation, is that even if only important features are preserved, the resulting reduced space is not suitable for the construction of an accurate mapping - perhaps DR results in non-continuous disjoint reduced input subspaces which could be remedied with the construction of multiple local PCE surrogate models. 

\review{Finally, in Figure \ref{fig:AP3-time-evolution} we present the evolution of $v(t,x,y)$ for 5 time steps $t=(0,0.25,0.5,0.75,1)$. The reference response is shown in the first row, the LLE PCE (optimal surrogate for $(\alpha_1^{(1)}, \alpha_2^{(1)})$) in the second row, and the point-wise relative error in the third row, for each one of the three test cases (a,b,c).} We observe that early in the simulation, the relative error is on the order of $10^{-4}$, while for the last time step the error is higher, although still within acceptable bounds. We note that the current application allowed us to identify certain limitations of the proposed approach. Particularly, the m-PCE accuracy diminishes when input random fields have high local variability (hence very large intrinsic dimension) and also when the solution exhibits strongly nonlinear or discontinuous behavior. We anticipate that in such cases, more expressive surrogate techniques such as DNNs could be employed and result in better accuracy.

\subsection{Overall method comparison}
\label{overall}

\begin{figure}[ht!]
\begin{center}
\includegraphics[width=0.7\textwidth]{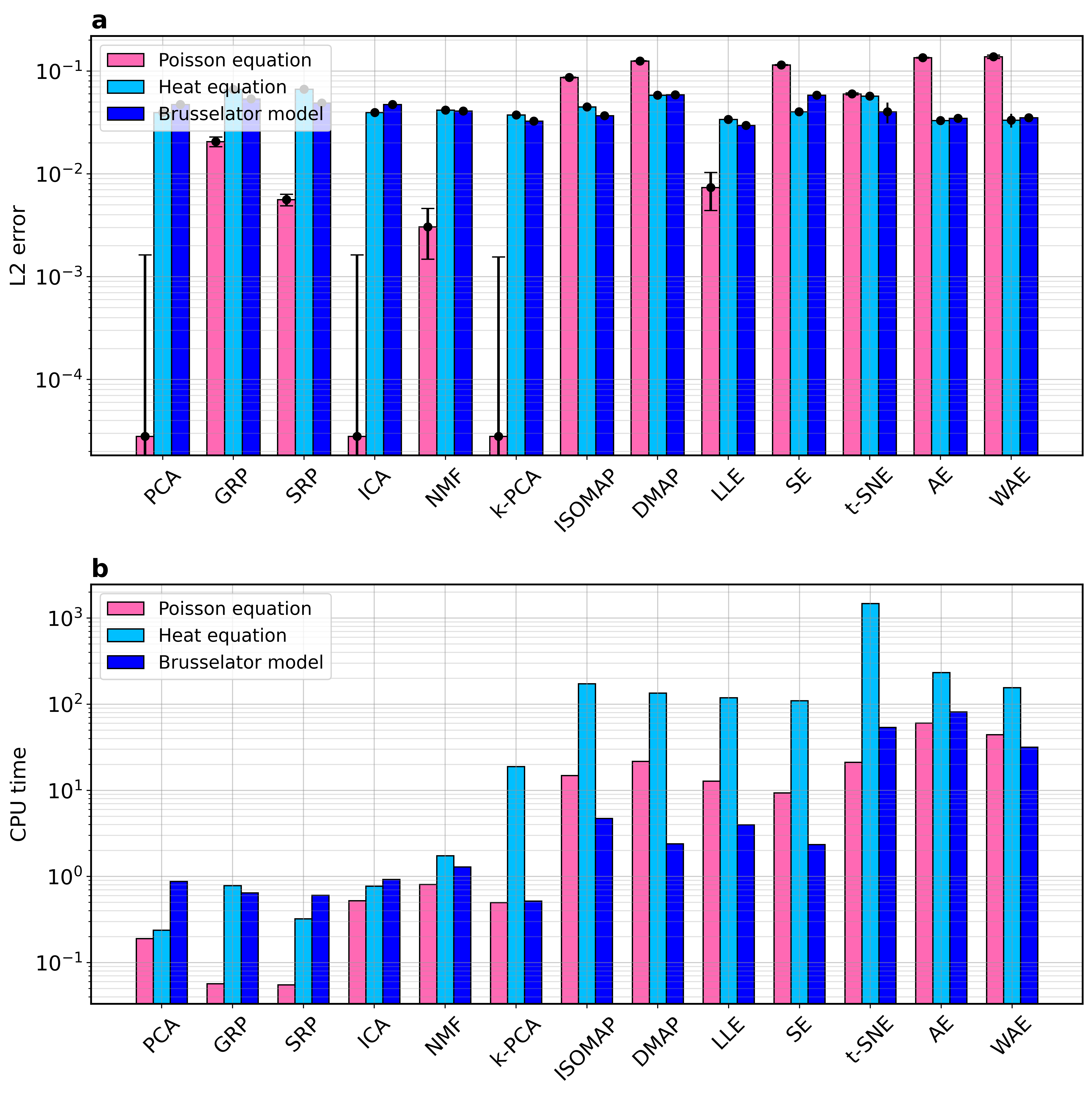}
\caption{\review{Overall comparison between the thirteen DR methods in terms of (a) relative error (with error bars corresponding to one standard deviation) and (b) CPU time of surrogate training. Pink bars represent the Poisson equation, blue bars the heat equation and dark blue bars the Brusselator model}}
\label{fig:overall-comparison}
\end{center}
\end{figure}

In this Section, we compare the performance of all DR methods in terms of the predictive ability of the PCE surrogate. In Figure \ref{fig:overall-comparison}, we present the (a) average relative $L_2$ error for all thirteen DR methods and (b) the CPU time in two bar graphs. As discussed in previous sections, simple methods such as PCA and ICA performed best for the 1D Poisson equation when considering both CPU time and surrogate accuracy. This is expected given the relative simplicity of the random field expansion. For more complex problems (complexity stems from both the nature of input stochastic fields or/and the model itself) like the heat equation or the Brusselator model, nonlinear methods such as AE or LLE performed very well in reducing the stochastic input dimension for accurate PCE. One important finding is that, when CPU time is of importance, simpler methods such as PCA, k-PCA and ICA were able to result in both good accuracy and fast training. Figure \ref{fig:overall-comparison}(b), clearly shows an upward trend in CPU time as the DR method gets more complicated (nonlinear, non-convex) that can result in up to 4 orders of magnitude higher expense compared to simple methods. A key takeaway from this comparative study is to start simple. During the past few years there has been a trend in employing overly complex and overparameterized techniques (e.g., DNNs) to approximate ODEs and PDEs that could simply be emulated with simpler methods as presented in this work. As shown, such methods can be employed both for predictive analysis and probabilistic modeling (uncertainty propagation, moment estimation). One should resort to more advanced methods only when complexity of the system necessitates and simple methods fail to result in satisfactory predictive accuracy.

\section{Discussion}
\label{S:Discussion}

Dimensionality reduction, an unsupervised learning task, has been established as a powerful pre-processing tool in cases where the high dimensionality of data would otherwise lead to burdensome computations. Within the context of high-dimensional UQ, only a few methods have been employed for the projection of inputs onto lower-dimensional embeddings for the construction of surrogates that enable inexpensive MC simulations and therefore UQ tasks, such as moment estimation and uncertainty propagation. Furthermore, the appropriateness of each method for various types of physics-based problems has not been investigated. In this work, we explored the functionalities, advantages and limitations associated with various DR methods applied for high-dimensional UQ in black-box-type models. The process of mapping a set of high-dimensional data to a low-dimensional compressed representation possesses numerous challenges. The selection of a suitable method according to the type of data is the most important issue that needs to be addressed. 

In this work, the methods used to generate high-dimensional input stochastic fields (KLE, SRM) result in both compact and less compact representations, allowing the discovery of a latent space with low-to-medium dimensionality.
The comparative analysis revealed that, for simple problems simulated on 1D domains, linear methods outperformed the more sophisticated nonlinear ones in terms of prediction accuracy, with the additional advantage of a very fast training. For more complicated problems, e.g. ones with input stochastic fields of varying lengthscales where the goal is to predict the model response in both space and time, nonlinear methods outperformed linear ones, with the downside of being more CPU-intensive, in particular exceeding the cost of linear methods from 1 to 3 orders of magnitude. In most cases, the optimal surrogate is constructed with a very small maximum polynomial degree $(s_{max}=2-4)$, indicating the ability of DR methods to reveal a smooth underlying embedding which can be mapped to responses via a lower-order polynomial expansion. Interestingly, when both predictive accuracy and CPU time are considered, simple methods such as PCA, k-PCA and ICA often outperformed the more complex and sophisticated approaches. Importantly, our results indicate that even for complex physics-based applications with high-dimensional input uncertainties, the m-PCE approach can be employed with simple DR algorithms and achieve comparable predictive accuracy to those of much more expensive surrogate modeling approaches proposed in the literature. For example, a combination of a standard PCA for DR of stochastic inputs and PCE for surrogate modeling performs as well as the use of expensive DNNs used for both the DR and surrogate modeling tasks proposed in \cite{tripathy2018deep}, but with a training time in the order of seconds (see the test case investigated in Section \ref{AP3}). Within the context of physics-based modeling and UQ, we showed that the m-PCE approach can provide a cost-effective solution for surrogate modeling and propagation of uncertainties in both simple and complex physics-based problems. 

We found that the predictive ability of the surrogate is based on both the complexity of the input-output mapping and the existence of a low-dimensional embedding of input uncertain data. Admittedly, there will be cases where high-dimensional inputs do not possess a low-dimensional structure that can be revealed with a DR method. In cases where the intrinsic dimensionality of input data is very high (e.g., for stochastic fields with small lengthscales generated with the spectral representation method, see example in Section \ref{AP2}), the construction of a PCE surrogate may be intractable. On the other hand, if a number of critical features are neglected to achieve a lower input dimensionality, the resulting embedding will not preserve the variance and local structure of the original data which will lead to the construction of an inefficient mapping between inputs and QoIs. These limitations form challenges to be addressed in future work. We anticipate that this study will assist researchers in the process of selecting suitable DR techniques for enhancing predictive analysis and UQ and alleviate the curse of dimensionality in complex problems associated with high-dimensional uncertainties. 

\section{Acknowledgements}
This material is based upon work supported by the U.S. Department of Energy, Office of Science, Office of Advanced Scientific Computing Research under Award Number DE-SC0020428. 
D. Loukrezis acknowledges the support of the Graduate School Computational Engineering within the Centre for Computational Engineering at the Technische Universit\"at Darmstadt.

\bibliography{sample}

\end{document}